\pdfoutput=1

\documentclass[sigconf]{acmart}

\usepackage{booktabs}
\usepackage{caption}
\usepackage{setspace}
\usepackage{enumitem}

\usepackage{subcaption}
\usepackage{hyperref}
\usepackage{bbm}
\usepackage{algorithm}
\usepackage{algorithmic}
\usepackage{tabularx}
\usepackage{multirow}

\usepackage{booktabs,subcaption,amsfonts,dcolumn}
\newcolumntype{d}[1]{D..{#1}}

\usepackage{diagbox}
\usepackage{flushend}

\usepackage{mathtools}

\def \eg {\emph{e.g.}, }
\def \ie {\emph{i.e.}, }

\copyrightyear{2019}
\acmYear{2019} 
\setcopyright{iw3c2w3}
\acmConference[WWW '19]{Proceedings of the 2019 World Wide Web Conference}{May 13--17, 2019}{San Francisco, CA, USA}
\acmBooktitle{Proceedings of the 2019 World Wide Web Conference (WWW '19), May 13--17, 2019, San Francisco, CA, USA}
\acmPrice{}
\acmDOI{10.1145/3308558.3313426}
\acmISBN{978-1-4503-6674-8/19/05}

\begin{document}

\title[Learning Sensing Signals with Short-Time Fourier Neural Network]{STFNets: Learning Sensing Signals from the Time-Frequency Perspective with Short-Time Fourier Neural Networks}

\author{Shuochao Yao$^1$, Ailing Piao$^2$, Wenjun Jiang$^3$, Yiran Zhao$^1$, Huajie Shao$^1$, Shengzhong Liu$^1$, Dongxin Liu$^1$, Jinyang Li$^1$, Tianshi Wang$^1$, Shaohan Hu$^4$, Lu Su$^3$, Jiawei Han$^1$, Tarek Abdelzaher$^1$}
\affiliation{%
  \institution{$^1$University of Illinois at Urbana-Champaign
  $^2$University of Washington 
  $^3$State University of New York at Buffalo
  $^4$IBM
  }
\institution{$^1$\{syao9, zhao97, hshao5, sl29, dongxin3, jinyang7, tianshi3, hanj, zaher\}@illinois.edu\\
$^2$alpiao@uw.edu 
$^3$\{wenjunji, lusu\}@buffalo.edu
$^4$shaohan.hu@ibm.com
}
  }

\sloppy

\begin{abstract}
Recent advances in deep learning motivate the use of deep neural networks in Internet-of-Things (IoT) applications. 
These networks are modelled after signal processing in the human brain, thereby leading to significant advantages at perceptual tasks such as vision and speech recognition. IoT applications, however, often measure {\em physical phenomena\/}, where the underlying physics (such as inertia, wireless signal propagation, or the natural frequency of oscillation) are fundamentally a function of signal frequencies, offering better features in the {\em frequency domain\/}. This observation leads to a fundamental question: For IoT applications, can one develop a new brand of neural network structures that synthesize features inspired not only by the biology of human perception but also by the fundamental nature of physics? Hence, in this paper, instead of using conventional building blocks (\eg convolutional and recurrent layers), we propose a new foundational neural network building block, the Short-Time Fourier Neural Network (STFNet). It integrates a widely-used time-frequency analysis method, the Short-Time Fourier Transform, into data processing to learn features directly in the frequency domain, where the physics of underlying phenomena leave better footprints. STFNets bring additional flexibility to time-frequency analysis by offering novel nonlinear learnable operations that are spectral-compatible. Moreover, STFNets show that transforming signals to a domain that is more connected to the underlying physics greatly simplifies the learning process. We demonstrate the effectiveness of STFNets with extensive experiments on a wide range of sensing inputs, including motion sensors, WiFi, ultrasound, and visible light. STFNets significantly outperform the state-of-the-art deep learning models in all experiments. A STFNet, therefore, demonstrates superior capability as the fundamental building block of deep neural networks for IoT applications for various sensor inputs.
\end{abstract}



\renewcommand{\authors}{Shuochao Yao, Ailing Piao, Wenjun Jiang, Yiran Zhao, Huajie Shao, Shengzhong Liu, Dongxin Liu, Jinyang Li, Tianshi Wang, Shaohan Hu, Lu Su, Jiawei Han, Tarek Abdelzaher}

\maketitle

\renewcommand{\shortauthors}{S. Yao et al.}

{

\section{Introduction}
Motivated by the needs of IoT applications, this paper presents a principled way of designing deep neural networks that learn (from IoT sensing signals) features inspired by the fundamental properties of the underlying domain of measurements; namely, properties of physical signals. We refer by IoT applications to those where sensors measure some physical quantities, generating (possibly complex and multi-dimentional) time-series data, typically reflecting some underlying physical process. The human brain (whose wiring inspires the structure of conventional neural networks) extracts features well-suited for {\em external perceptual\/} tasks, which explains the great success of such networks at those tasks. In contrast, the internal {\em physical processes\/} underlying sensor measurements in IoT systems have properties (such as physical intertia, characteristics of wireless signal propagation, and signal resonance) that depend more on signal {\em frequency\/}, motivating feature extraction in the {\em frequency domain\/}. It is no coincidence that much of classical signal processing literature works by transforming time-series data to the frequency domain first. To help capture signatures of internal physical processes the way a brain captures their externally perceived properties, this paper develops a new neural network block designed specifically for learning in the frequency domain.

The design of neural network structures greatly influences efficiency of signal modelling and ease of extraction of hidden patterns. Convolutional neural networks (CNNs) for image recognition, for example, align perfectly with biological studies of the visual cortex~\cite{hubel1968receptive} and with domain knowledge in digital image processing~\cite{gonzalez2002digital}. We thus ask a fundamental question: what structures are well-suited for the domain of physical sensor measurements, which we henceforth call the domain of IoT? 

Previous research on customizing deep learning models to the needs of IoT applications~\cite{yao2018deep,lane2015deepear,yao2017deepsense} mainly focused on designing neural network structures that integrate conventional deep learning components, such as convolutional and recurrent layers, to extract spatial and temporal properties of inputs. On the other hand, since the physics of measured phenomena are best expressed in the frequency domain, decades of research on signal processing developed powerful techniques for time-frequency analysis of signals, including motion sensor signals~\cite{stisen2015smart, hemminki2013accelerometer}, radio frequency signals~\cite{wang2015understanding,pu2013whole}, acoustic signals~\cite{gupta2012soundwave,chen2014airlink}, and visible light signals~\cite{li2016practical}. A popular transform that maps time-series measurements to the frequency domain is the  
Short-Time Fourier Transform (STFT). We, therefore, propose a new neural network model, namely, Short-Time Fourier Neural Networks (STFNets) that operate directly in the frequency domain. 

One potential approach for learning in the frequency domain might simply be to convert sensing signals into the frequency domain first, and then apply conventional neural network components, possibly extending them to support operations on complex-numbers so they can represent frequency-domain quantities~\cite{trabelsi2017deep}. These approaches miss two key opportunities for improvement, described below, that we take advantage of in this work. As a result, our work leads to more accurate results, as shown in the evaluation section.  The two reasons that account for our improvements are as follows.

First, different from traditional neural networks, where the internal representations constitute features with no physical meaning, the internal representations in STFNet leverage frequency domain semantics that encode time and frequency information.
All operations and learnable parameters we propose are explicitly made compatible with the basic properties of spectral data, and align corresponding frequency and time components. In our design, we categorize spectral manipulations into three main types: filtering, convolution, and pooling. Filtering refers to the general spectral filtering and global template matching operation; convolution refers to the local motif detection including shift detection and local template detection; and pooling refers to dimension reduction over the frequency domain. We then design the spectral-compatible parameters and operating rules for these three manipulation categories respectively, which have shown superior performance in evaluations compared to the application of {\em conventional\/} neural networks in the domain of complex-numbers.

Second, transforming signals to the frequency domain is governed by the {\em uncertainty principle\/}~\cite{smith2007mathematics}. The transformed representation cannot achieve both a high frequency resolution and a high time resolution at the same time. In STFT, the time-frequency resolution is controlled by the length of the sliding window (the length of the part of the time-series being converted at a time). With a longer window, we can obtain a finer-grained frequency representation. However, we then cannot achieve a time resolution smaller than the window size. The uncertainty principle causes a dilemma in traditional time-frequency analysis. One often needs to guess the best time-frequency resolution using trial and error. In STFNet, we circumvent this dilemma by simultaneously computing multiple STFTs with different time-frequency resolutions. The representations with different time-frequency resolutions are then mutually enhanced in a data-driven manner. The network then automatically learns the best resolution or resolutions, where the most useful features are present.
STFNet defines a formal way to extract features from multiple time-frequency transformations with the same set of spectral-compatible operations and parameters, which greatly reduces model complexity while improving accuracy.

We demonstrate the effectiveness of STFNet through extensive experiments with various sensing modalities. During the evaluation, we focus on device-based and device-free human activity recognition with a broad range of sensing modalities, including motion sensors (accelerometer and gyroscopes), WiFi, ultrasound, and visible light. The experimental results validate the design settings of STFNets and illustrate their superior accuracy compared to the state-of-the-art deep learning frameworks for IoT applications.

Broadly speaking, the main contributions of this paper to the general research landscape of deep learning and IoT are twofold:
\begin{enumerate}
\item STFNet presents a principled way of designing neural networks that reveal the key properties of physical processes underlying the sensing signals from the time-frequency perspective.
\item STFNet unveils the benefit of incorporating domain-specific analytic modelling and transformation techniques into the neural network design.
\end{enumerate}

The rest of paper is organized as follows. Section~\ref{sec:related} introduces related work on deep learning in the context of mobile sensing as well as deep learning for spectral representations. We introduce the detailed technical design of STFNet in Section~\ref{sec:model}. The evaluation is presented in Section~\ref{sec:evaluation}. Finally, we discuss the results in Section~\ref{sec:discussion} and conclude in Section~\ref{sec:conclusion}.

\vspace{-0.35cm}
\section{Related Work}~\label{sec:related}
The impressive achievements in image classification using deep neural networks at the turn of the decade~\cite{krizhevsky2012imagenet} precipitated a re-emergence of interest in deep learning. Deep neural networks have achieved significant accuracy improvements in a broad spectrum of areas, including computer vision~\cite{simonyan2014very,he2016deep}, natural language processing~\cite{collobert2011natural,bahdanau2014neural}, and network analysis~\cite{perozzi2014deepwalk,kipf2016semi}.


Recent efforts applied deep learning in the context of IoT. 
In order to improve the predictive accuracy of IoT applications, researchers employed deep learning to model complicated sensing tasks~\cite{lane2015deepear,yao2017deepsense}. In order to improve system efficiency at executing neural networks on low-end IoT devices, efforts have been made to compress model parameters and/or structures in a manner that does not entail (almost any) accuracy loss~\cite{yao2017deepiot,yao2018fastdeepiot,bhattacharya2016sparsification,han2015deep}. 
Recent work in the context of IoT also addressed mathematical foundations for quantifying confidence in deep learning predictions to support mission-critical applications. The work produced deep neural networks that offer well-calibrated uncertainty estimates in results~\cite{yao2018rdeepsense,yao2018apdeepsense,gal2015dropout,gal2016theoretically}.
Finally, the challenge of insufficient labeling of IoT data was addressed by introducing semi-supervised approaches for deep learning that allow neural network training using mostly {\em unlabeled\/} data~\cite{yao2018sensegan}.
However, none of the aforementioned IoT-inspired efforts addressed the customization of learning machinery to a different signal space inspired by the physics of measured processes; namely, the frequency domain.

\begin{figure}[htb]
\vspace{-0.3cm}
\centering
\includegraphics[width=0.7\linewidth]{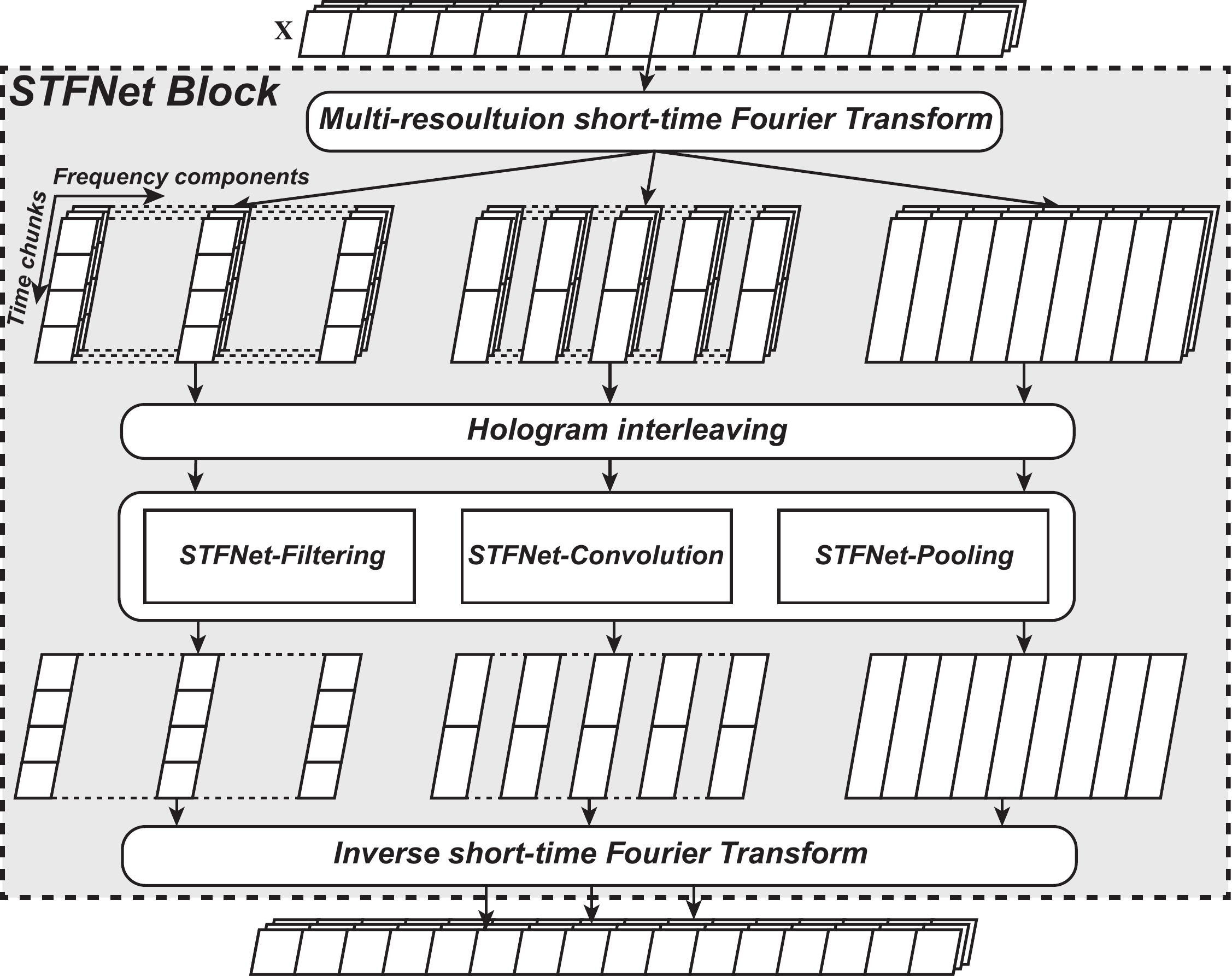}
\vspace{-0.3cm}
\caption{The overview design of STFNet block.}
\label{fig:STFNet_overview}
\end{figure}
\setlength{\textfloatsep}{0pt}

To fill the above gap, recent work in machine learning focused on extending deep neural networks to complex numbers and spectral representations. 
Trabelsi et al. propose deep complex networks, investigating the complex-value neural network structure~\cite{trabelsi2017deep}. However, they mainly concentrate on the problems of initialization, normalization, and activation functions when extending real-valued operations directly into the complex-value domain. Their designs focus more on complex-value representations than spectral representations, and do not take the properties of spectral data into consideration. 
Rippel et al. study spectral representations for convolutional neural networks~\cite{rippel2015spectral}. However, their study focuses on spectral parametrizing of standard CNNs, instead of designing operations customized for spectral data. In addition, their work treats input data fully from the frequency perspective instead of the time-frequency perspective. 
Yao et al. propose a neural network that takes short-time Fourier transformed data as inputs~\cite{yao2017deepsense}. Yet their design uses traditional CNNs and RNNs, combining the real and imagery parts of complex-value inputs as additional features.

To the best of our knowledge, STFNet is the first work that integrates neural networks with traditional time-frequency analysis, and designs fundamental spectral-compatible operations for Fourier-transformed representations. Our study shows that the approach leads to improved accuracy compared to the state of the art. It implies that integrating neural networks with domain-inspired transformation techniques (in our case, the Fourier Transform of physical time-series signals) projects input signals into a space that significantly facilitates the learning process.

\section{Short-Time Fourier Neural Networks}~\label{sec:model}
We introduce the technical details of STFNets in this section. We separate the technical descriptions into six parts. In the first two subsections, we provide some background followed by a high-level overview of STFNet components, including (i) hologram interleaving, (ii) STFNet-filtering, (iii) STFNet-convolution, and (iv) STFNet-pooling. In the remaining four subsections, we describe the technical details of each of these components, respectively.

\subsection{Background and STFNet Overview}
IoT devices sample the physical environment generating time-series data. Discrete Fourier Transform (DFT) is a mathematical tool that converts $n$ samples over time (with a sampling rate of $f_s$) into a $n$ components in frequency (with a frequency step of $f_s/n$). The more samples are selected, the finer the component resolution is in frequency. 
\begin{figure}[htb]
\vspace{-0.3cm}
\centering
\includegraphics[width=0.75\linewidth]{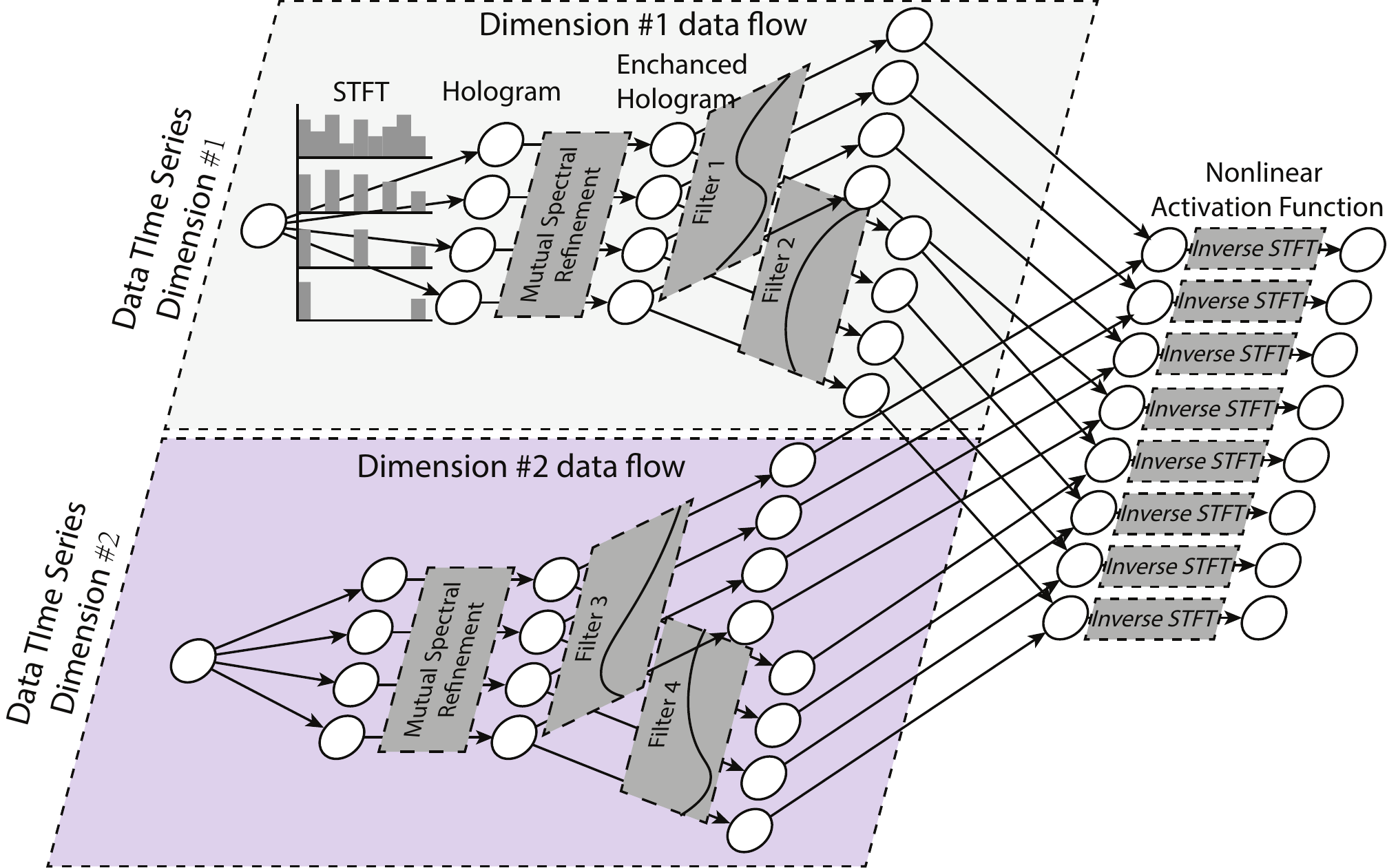}
\vspace{-0.3cm}
\caption{Data Flow within a block of STFNet.}
\label{fig:dataflow}
\end{figure}
We can always transform the whole sequence of data with DFT, achieving a high frequency resolution. However, we then lose information on signal evolution over time, or the time resolution. In order to solve this problem, Short-Time Fourier Transform (STFT) divides a longer time signal into shorter segments of equal length and computes DTF separately on each shorter segment. By losing a certain degree of frequency resolution, STFT helps us regain the time resolution to some extent. In choosing $n$, there arises a fundamental trade-off between the attainable time and frequency resolution, which is called the {\em uncertainty principle\/}~\cite{smith2007mathematics}. 
For the purposes of learning to pedict a given output, the optimal trade-off point depends on the time and frequency granularity of the features that best determine the outputs we want to reproduce. The goal of STFNets is thus to learn frequency domain features that predict the output, while at the same time learn the best resolution trade-off point in which the relevant features exist.

The building component of an STFNet is an {\em STFNet block\/}, shown in Figure~\ref{fig:STFNet_overview}. An STFNet block is the layer-equivalent in our neural network. The larger network would normally be composed by stacking such layers. Within each block, STFNet circumvents the uncertainty principle by computing multiple STFT representations with different time-frequency resolutions. Collectively, these representations constitute what we call the {\em time-frequency hologram\/}. And we call an individual time-frequency signal representation, a hologram 
representation. They are then used to mutually enhance each other by filling-in missing frequency components in each.

Candidate frequency-domain features are then extracted from these enhanced representations via general spectral manipulations that come in two flavors; filtering and convolution. They represent global and local feature extraction operations, respectively. The filtering and convolution kernels are learnable, making each STFNet layer a building block for spectral manipulation and learnable frequency domain feature extraction. In addition, we also design a new mechanism, called pooling, for frequency domain dimensionality reduction in STFNets. Combinations of features extracted using the above manipulations then pass through activation functions and an inverse STFT transform to produce (filtered) outputs in the time domain. Stacking STFNet blocks has the effect of producing progressively sharper (i.e., higher order) filters to shape the frequency domain signal representation into more relevant and more fine-tuned features.

Figure~\ref{fig:dataflow} gives an example of an SFTNet block that accepts as input a two-dimensional time-series signal (e.g., 2D accelerometers data). Each dimension is then transformed to the frequency domain at four different resolutions using STFT, generating four different internal nodes, each representing the signal in the frequency domain at a different time-frequency resolution. Collectively, the four representations constitute the hologram. In the next step, mutual enhancements are done improving all representations. Each representation then undergoes a variety of alternative spectral manipulations (called ``filters" in the figure). Two filters are shown in the figure for each dimension. The parameters of these filters are the weights multiplied by the frequency components of the filter input; a different weight per component. These parameters are what the network learns. Note that, a filter does not change the time-frequency resolution of the corresponding input. Filter outputs of the same time-frequency resolution are then combined additively across all dimensions and passed through a non-linear activation function (as in a conventional convolutional neural network). An inverse STFT brings each such combined output back to the time domain, where it becomes an input to the next STFNet block. (Alternatively, the inverse STFT can be applied after dimension combination and before the activation function.) Hence, each output time-series is produced by applying spectral manipulation and fusion to one particular time-frequency resolution of all input time-series. Once converted to the time domain, however, the output time-series can be resampled in the next block at different time-frequency resolutions again. The goal of STFNet is to learn the weighting of different frequency components within each filter in each block such that features are produced that best predict final network outputs.  

\subsection{STFNet Block Fundamentals}~\label{sec:STFNet_overview}
In this subsection, we introduce the formulation of our design elements wihin each STFNet block. In the rest of this paper, all vectors are denoted by bold lower-case letters (e.g., $\mathbf{x}$ and $\mathbf{y}$), while matrices and tensors are represented by bold upper-case letters (e.g., $\mathbf{X}$ and $\mathbf{Y}$). For a vector $\mathbf{x}$, the $j^{th}$ element is denoted by $\mathbf{x}_{[j]}$. For a tensor $\mathbf{X}$, the $t^{th}$ matrix along the third axis is denoted by $\mathbf{X}_{[\cdot, \cdot, t]}$, and other slicing denotations are defined similarly. We use calligraphic letters to denote sets (e.g., $\mathcal{X}$ and $\mathcal{Y}$). For set $\mathcal{X}$, $|\mathcal{X}|$ denotes the cardinality.

We denote the input to the STFNet block as $\mathbf{X}\in \mathbb{R}^{T \times D}$, 
where we divide the input $D$-dimension time-series into windows of size $T$ samples. We call $T$ the signal length and $D$ the signal dimension.
Since we concentrate on sensing signals, we assume that all the raw and internal-manipulated sensing signals are real-valued in time domain. 

As shown in Figure~\ref{fig:STFNet_overview}, the input signal $\mathbf{X}$ first goes through a multi-resolution short-time Fourier transform ($\text{Multi\_STFT}$), which is a compound traditional short-time Fourier transform ($\text{STFT}$), to provide a time-frequency hologram of the signal. $\text{STFT}$ breaks the original signal up into chunks with a sliding window, where sliding window $\mathbf{W}(t)$ with width $\tau$ only has non-zero values for $1 \le t\le \tau$. Then each chunk is Discrete-Fourier transformed, 
\begin{equation}
\small
\mathbf{STFT}^{(\tau, s)}(\mathbf{X})_{[m, k, d]}  = \sum_{t=1}^T \mathbf{X}_{[t, d]} \cdot \mathbf{W}(t - s\cdot m) \cdot e^{-j\frac{2\pi k}{\tau} (t - s\cdot m)},
\label{eqn:STFT}
\end{equation}
where $\mathbf{STFT}^{(\tau, s)}(\mathbf{X}) \in \mathbb{C}^{M\times K\times D}$ denotes the short-time Fourier transform with width $\tau$ and sliding step $s$. $M$ denotes the number of time chunks. $K$ denotes the number of frequency components. Since input signal $\mathbf{X}$ is real-valued, its discrete Fourier transform is conjugate symmetric. Therefore, we only need the $\lfloor\tau/2\rfloor + 1$ frequency components to represent the signal, \ie $K = \lfloor\tau/2\rfloor + 1$.
In this paper, we focus on sliding chunks with rectangular window and no overlaps to simplify the formulation, \ie $s = \tau$ and $M = T/\tau$. We therefore denote of short-time Fourier transform as  $\mathbf{STFT}^{(\tau)}(\mathbf{X})$.

\begin{figure}[!htb]
\vspace{-0.3cm}
\centering
\includegraphics[width=0.7\linewidth]{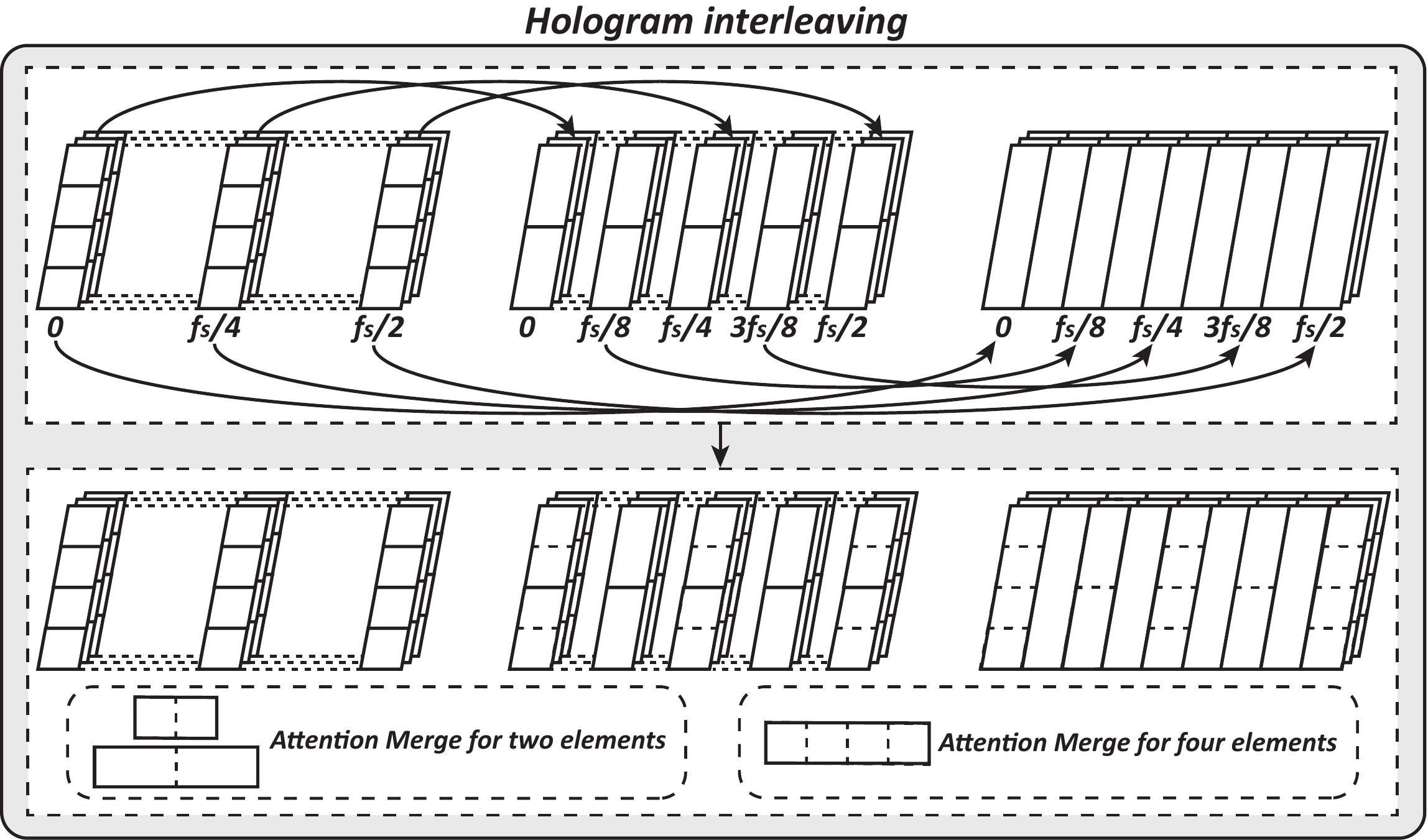}
\vspace{-0.3cm}
\caption{The design of hologram interleaving.}
\label{fig:hologram_interleaving}
\vspace{-0.1cm}
\end{figure}

The $\text{Multi\_STFT}$ operation is composed of multiple short-time Fourier transform with different window widths $\mathcal{T} = \{\tau_i\}$. The window width, $\tau_i$, determines the time-frequency resolution of $\text{STFT}$. Larger $\tau_i$ provides better frequency resolution, while smaller $\tau_i$ provides better time resolution. In this paper, we set the window widths to be powers of $2$, \ie $\tau_i = 2^{p_i}$ $\forall p_i \in \mathbb{Z}_0^+$, to simplify the design later. We can thus formulate $\text{Multi\_STFT}$ as:
\begin{equation}
\small
\begin{split}
\mathbf{Multi\_STFT}^{(\mathcal{T})}\{\mathbf{X}\} & = \big\{\mathbf{STFT}^{(\tau_i)}(\mathbf{X})\big\} \text{ for }2^{p_i} \in \mathcal{T}.
\end{split}
\label{eqn:Multi_STFT}
\end{equation}

Next, according to Figure~\ref{fig:STFNet_overview}, the multi-resolution representations go into the hologram interleaving component, which 
enables the representations to compensate and balance their time-frequency resolutions with each other. The technical details of the hologram interleaving component are introduced in Section~\ref{sec:STFNet_hologram}.

The STFNet block then manipulates multiple hologram representations with the same set of spectral-compatible operation(s), including STFNet-filtering, STFNet-convolution, and STFNet-pooling. We will formulate these operations in Section~\ref{sec:STFNet_filtering},~\ref{sec:STFNet_convolution}, and~\ref{sec:STFNet_pooling}, respectively.

Finally, the STFNet block converts the manipulated frequency representations back into the time domain with the inverse short-time Fourier transform. The resulting representations from different views of the hologram are weighted and merged as the input ``signal" for the next block. Since we merge the output representations from different views of the hologram, we reduce the output feature dimension of STFNet-filtering and convolution operations by the factor of $1/|\mathcal{T}|$ to prevent the dimension explosion.

\subsection{STFNet Hologram Interleaving}~\label{sec:STFNet_hologram}
In this subsection, we introduce the formulation of hologram interleaving. 
Due to the Fourier uncertainty principle, the representations in time-frequency hologram either have high time resolution or high frequency resolution. 
The hologram interleaving tries to use representations with high time resolution to instruct the representations with low time resolution to highlight the important components over time.
This is done by two steps:
\begin{enumerate}
\item Revealing the mathematical relationship of aligned time-frequency components among different representations in the time-frequency hologram.
\item Updating the original relationship in a data-driven manner through neural-network attention components.
\end{enumerate}

We start from the definition of time-frequency hologram, generated by $\text{Multi\_STFT}$ defined in~\eqref{eqn:Multi_STFT}. Note that, the window width set $\mathcal{T}$ is defined as $ \{2^{p_i}\}$, $ \forall p_i \in \mathbf{Z}_0^{+}$. 
Without loss of generality, an illustration of multi-resolution short-time Fourier transformed representations with input signal having length $16$ and signal dimension $3$ as well as $\mathcal{T} = \{4, 8, 16\}$ are illustrated in Figure~\ref{fig:hologram_interleaving}.

In order to find out the relationship of aligned time-frequency components, we start with the frequency-component dimension. Since different representations only change the window width $\tau_i$ of STFT but not the sampling frequency $f_s$ of input signal, these frequency components represent frequencies from $0$ to $f_s/2$ (Nyquist frequency) with step $f_s/\tau_i$. Then we can first obtain the relationship of frequency ranging steps among different representations, 
\vspace{-0.15cm}
\begin{equation}
\small
\forall p_i > p_j \textrm{ , } \frac{f_s/\tau_j}{f_s/\tau_i} = 2^{p_i - p_j} \in \mathbf{Z}_0^{+}.
\vspace{-0.15cm}
\label{eqn:freq_align}
\end{equation}

Therefore, a low frequency-resolution representation (with window width $2^{p_j}$) can find their frequency-equivalent  counterparts for every $2^{p_i - p_j}$ frequency components in a high frequency-resolution representation (with window width $2^{p_i}$). The upper part of Figure~\ref{fig:hologram_interleaving} provides a simple illustration of such relationship. In the following analysis, we will use the original index $k$ and corresponding frequency $k\cdot f_s/\tau_i$ interchangeably to recall the frequency component from the time-frequency hologram $\mathbf{STFT}^{(\tau)}(\mathbf{X})_{[m, k, d]} $.

Next, we analyze the relationship over the time-chunk dimension, when two representations have frequency-equivalent components.
Note that time chunks in $\mathbf{STFT}^{(\tau)}(\mathbf{X})$ are generated by sliding rectangular window without overlap. Based on~\eqref{eqn:STFT}, for representations having window widths $\tau_i=2^{p_i}$ and $\tau_j=2^{p_j}$ ($p_i > p_j$),
\vspace{-0.3cm}
\begin{equation}
\small
\begin{split}
\mathbf{STFT}^{(\tau_i)}(\mathbf{X})_{[m, 2^{p_i - p_j}k, d]}  = \sum_{t=2^{p_i} m + 1}^{2^{p_i}(m+1)} \mathbf{X}_{[t, d]} \cdot e^{-j\frac{2\pi  2^{p_i - p_j} k}{2^{p_i}} (t - m\cdot 2^{p_i})}, \\
= \sum_{m_j = 2^{p_i-p_j}m}^{2^{p_i-p_j}(m+1) -1} \sum_{t=m_j +1}^{ 2^{p_j}(m_j+1)} \mathbf{X}_{[t, d]} \cdot e^{-j\frac{2\pi k}{2^{p_j}} (t - m\cdot 2^{p_j})},\\
= \sum_{m_j = 2^{p_i-p_j}m}^{2^{p_i-p_j}(m+1) -1} \mathbf{STFT}^{(\tau_j)}(\mathbf{X})_{[m_j, k, d]}. \phantom{aaaaaaaaaaa}
\end{split}
\label{eqn:time_align}
\end{equation}
\vspace{-0.3cm}

Therefore, given the equivalent frequency component, a time component in low time-resolution representation (with window width $2^{p_i}$) is the sum of $2^{p_i - p_j}$ aligned time components of the high time-resolution representation (with window width $2^{p_j}$). As a toy example in Figure~\ref{fig:hologram_interleaving}, the first row of the middle tensor is equal to the sum of first two rows of the left tensor for frequencies $0$, $f_s/4$, and $f_s/2$. The row of the right tensor is equal to the sum of four rows of the left tensor for frequencies $0$, $f_s/4$, and $f_s/2$. The row of the right tensor is equal to the sum of two rows of the middle tensor for frequencies $f_s/8$ and $3f_s/8$, etc.

According to the analysis above, the high frequency-resolution representations lose their fine-grained time resolutions at certain frequencies by summing the corresponding frequency components up over a range of time. However, the high time-resolution representations preserve these information. 

The idea of hologram interleaving is to replace the sum operation in high frequency-resolution representation with a weighted merge operation to highlight the important information over time. For a certain frequency component, the weight of merging is learnt through the most fine-grained information preserved 
in the time-frequency hologram. In this paper, we implement the weighted merge operation as a simple attention module. For a merging input $\mathbf{z} \in\mathbb{C}^{S \times 1}$, where $S$ is the number of elements to be merged, the merge operation is formulated as:

\vspace{-0.4cm}
\begin{equation}
\small
\begin{split}
\mathbf{a} &= \mathrm{softmax}( |\mathbf{W}_m  \mathbf{z}| ), \\
y &= S\times \mathbf{a}^\intercal  \mathbf{z},
\end{split}
\vspace{-0.2cm}
\label{eqn:time_align}
\end{equation}
\begin{figure}[!htb]
\vspace{-0.3cm}
\centering
\includegraphics[width=0.7\linewidth]{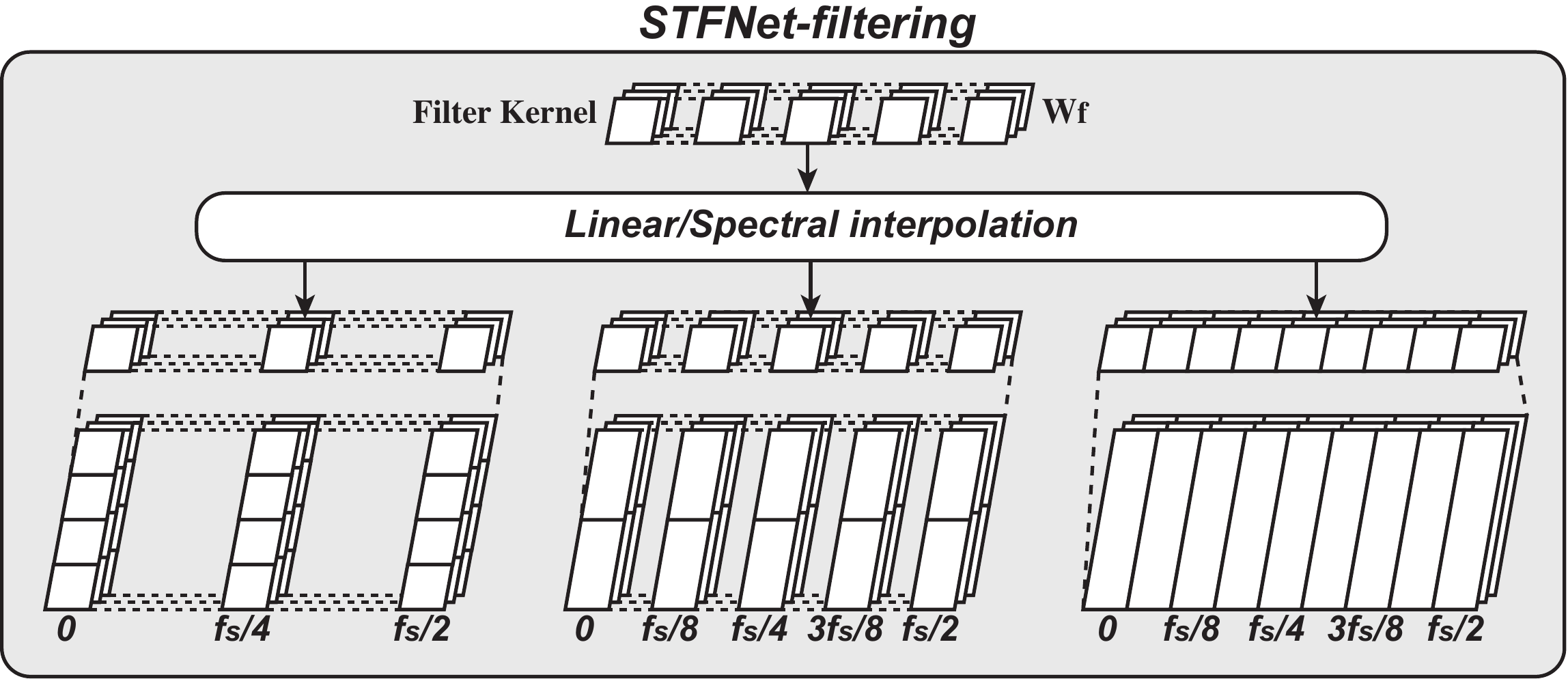}
\vspace{-0.3cm}
\caption{The STFNet-filtering operation.}
\label{fig:STFNet_filtering}
\end{figure}
where $|\cdot|$ is the piece-wise magnitude operation for complex-number vector; and $\mathbf{W}_m \in \mathbb{C}^{S\times S}$ is the learnable weight matrix. Notice that the final merged result is rescaled by the factor $S$ to imitate the ``sum" property of Fourier transform.

\subsection{STFNet-Filtering Operation}~\label{sec:STFNet_filtering}
Starting from this subsection, we will introduce our three spectral-compatible operations in STFNet. In each subsection, the introduction includes two main parts: 1) the basic formulation of proposed spectral-compatible operation, and 2) extending a single operation to multi-resolution data.

Spectral filtering is a widely-used operation in time-frequency analysis. The STFNet-filtering operation replaces the traditional manually designed spectral filter with a learnable weight that can update during the training process. 
Although the spectral filtering is equivalent to the time-domain convolution according to convolution theorem~\footnote{\url{https://en.wikipedia.org/wiki/Convolution_theorem}}, the filtering operation helps to handle the multi-resolution time-frequency analysis, and facilitates the parameterization and modelling. We denote the input tensor as $\mathbf{X} \in \mathbf{C}^{M\times K\times D}$, where $M$ is the number of time chunk, $K$ frequency component number, and $D$ input feature dimension. The STFNet-filtering operation is formulated as:
\begin{equation}
\small
\begin{split}
\mathbf{Y}_{[m, k, \cdot]}  = \mathbf{X}_{[m, k, \cdot]}  \mathbf{W}_{f[k, \cdot, \cdot]},
\end{split}
\label{eqn:STFNet_filtering}
\end{equation}
where $\mathbf{W}_f \in \mathbb{C}^{K\times D\times O}$ is the learnable weight matrix, $O$ the output feature dimension, and $\mathbf{Y}\in\mathbb{C}^{M\times K\times O}$ the output representation.

The function of STFNet-filtering operation is providing a set of learnable global frequency template matchings over the time. However, it is not straightforward to extend the matching operation to the representations with different time-frequency resolutions. Although we can create multiple $\mathbf{W}_f$ with different frequency resolutions $K$, it can introduce unnecessary complexity and redundancy.

STFNet-filtering solves this problem by interpolating the frequency components in weight matrix. As we mentioned in Section~\ref{sec:STFNet_hologram}, data in hologram with different frequency resolutions have the same frequency range (from $0$ to $f_s/2$) but different frequency steps ($f_s/\tau$). Therefore, STFNet-filtering operation only has one weight matrix $\mathbf{W}_f$ with $K = \lfloor\tau/2\rfloor + 1$ frequency components. When the operation input has $K' = \lfloor\tau'/2\rfloor + 1$ frequency components with $K' < K$, we can subsample the frequency components in $\mathbf{W}_f$.
When $K' > K$,  we interpolate the frequency components of $\mathbf{W}_f$. STFNet provides two kind of interpolation methods: 1) linear interpolation and 2) spectral interpolation.

\begin{figure}[!htb]
\vspace{-0.2cm}
\centering
\includegraphics[width=0.7\linewidth]{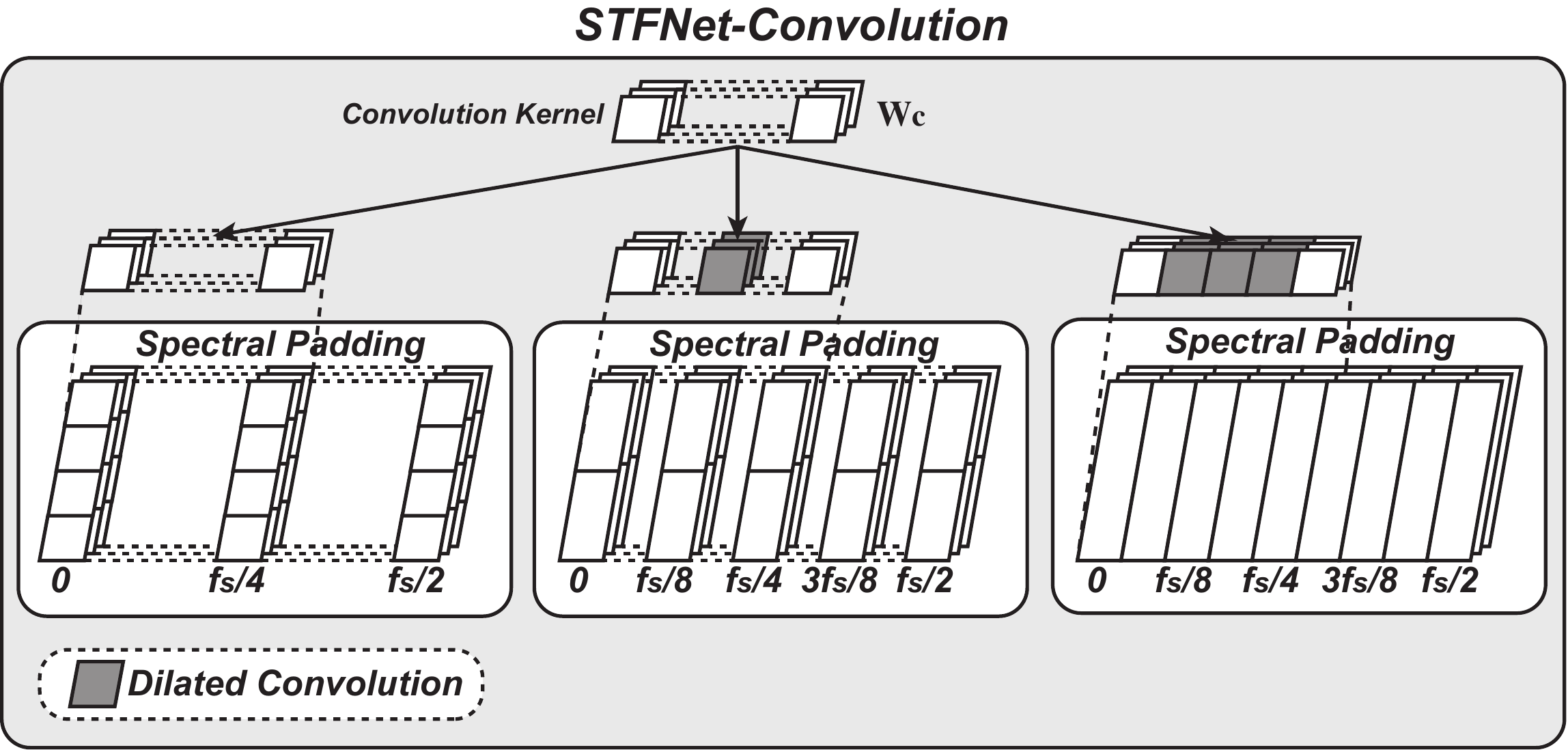}
\vspace{-0.3cm}
\caption{The STFNet-convolution operation with dilated configuration.}
\label{fig:STFNet_convolution}
\end{figure}

The linear interpolation generates the missing frequency components in extended weight matrix $\mathbf{W}'_{f} \in\mathbb{C}^{K'\times D\times O}$ from the two neighbouring frequency components in $\mathbf{W}_f$:
\begin{equation}
\small
\begin{split}
&k_l = \Big\lfloor k' \frac{\tau}{\tau'} \Big\rfloor \quad k_r = k_l +1, \\
&\mathbf{W}'_{f[k',\cdot, \cdot]} = \mathbf{W}_{f[k_l,\cdot, \cdot]} \Big(k_r - k'\frac{\tau}{\tau'}\Big) + \mathbf{W}_{f[k_r,\cdot, \cdot]} \Big( k'\frac{\tau}{\tau'} - k_l \Big).
\end{split}
\label{eqn:linear_interpolate}
\end{equation}

The spectral interpolation utilizes the relationship between discrete-time Fourier transform (DTFT) and discrete Fourier transform (DFT). For a time-limited signal (with length $\tau$), DTFT regards it as a infinite-length data with zeros outside the time-limited range, while DFT regards it as a $\tau$-periodic data. As a result, DTFT generates a continuous function over the frequency domain, while DFT generates a discrete function. 
Therefore, DFT can be regarded as a sampling of DTFT with step $f_s/\tau$. In order to increase the frequency resolution of $\mathbf{W}_f$, we can increase the sampling step from $f_s/\tau$ to $f_s/\tau'$, which is called spectral interpolation. Spectral interpolation can be done through zero padding in the time domain~\cite{smith2007mathematics},
\begin{equation}
\small
\begin{split}
\mathbf{W}'_{f[\cdot, d, o]} = \mathbf{DFT}\Big(\mathbf{ZeroPad}_{\tau' - \tau} \mathbf{IDFT}\big( \mathbf{W}_{f[\cdot, d, o]} \big) \Big),
\end{split}
\label{eqn:spectral_interpolate}
\end{equation}
where $\mathbf{ZeroPad}_t$ denotes padding $t$ zeros at the end of sequence, and $\mathbf{IDFT}(\cdot)$ denotes the inverse discrete Fourier transform. Please note that, if we pad infinite zeros to the IDFT result, then DFT turns into DTFT. An simple illustration of STFNet-filtering operation is shown in Figure~\ref{fig:STFNet_filtering}.

\subsection{STFNet-Convolution Operation}~\label{sec:STFNet_convolution}
In this subsection, we introduce our design of STFNet-convolution operation. Other than filtering operation that handles global pattern matching, we still need the convolution operation to deal with local motifs in the frequency domain.
We denote the input tensor as $\mathbf{X} \in \mathbf{C}^{M\times K\times D}$, where $M$ is the number of time chunk, $K$ number of frequency component, and $D$ input feature dimension.
The convolution operation involves two steps: 1) padding the input data, and 2) convolving with kernel weight matrix $\mathbf{W}_c \in \mathbb{C}^{1\times S \times D \times O}$, where $S$ is the kernel size along the frequency axis and $O$ is still the output feature dimension. 

Without the padding step, the output of convolution operation will shrink the number of frequency components, which may break the underlying structure and information in the frequency domain. Therefore, we need to pad extra ``frequency component" to keep the shape of output tensor unchanged compared to that of the input data. In the deep learning research, padding zeros is a common practice. Zero padding is reasonable for inputs such as images and signal in the time domain, meaning no additional information in the padding range. 
However, padding zero-valued frequency component introduces additional information in the frequency domain.

Therefore, STFNet-convolution operation proposes the spectral padding for time-frequency analysis.
According to the definition of DFT, transformed data is periodic within the frequency domain. In addition, if the original signal is real-valued, then the transformed data is conjugate symmetric within each period. 
Previously, we cut the number of frequency components of a $\tau$-length signal to $K = \lfloor\tau/2\rfloor +1$ for reducing the redundancy. In the spectral padding, we add these frequency components back according to the rule
\begin{figure}[!htb]
\vspace{-0.2cm}
\centering
\includegraphics[width=0.7\linewidth]{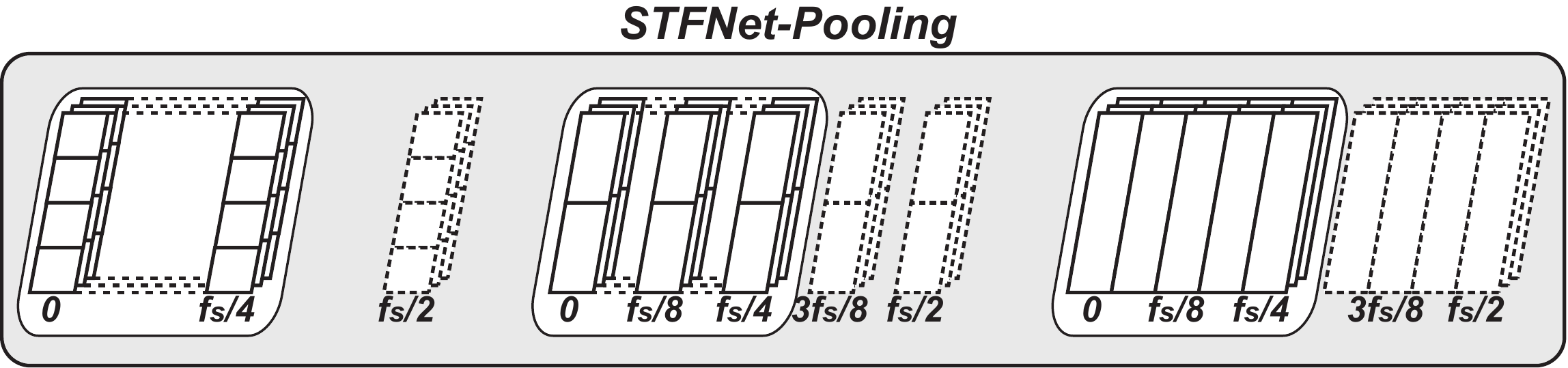}
\vspace{-0.3cm}
\caption{The low-pass STFNet-pooling operation.}
\label{fig:STFNet_pooling}
\end{figure}
\begin{equation}
\mathbf{X}_{[\cdot, \tau-k, \cdot]} = \mathbf{X}_{[\cdot,-k,  \cdot]} = \mathbf{X}_{[\cdot,k,  \cdot]}^*,
\label{eqn:spectral_padding}
\end{equation}
where $\mathbf{X}^*$ denotes complex conjugation. In addition, the number of padding before and after the input tensor is same as the previous padding techniques.

Then we can define the basic convolution operation in STFNet
\begin{equation}
\mathbf{Y} = \mathbf{SpectralPad}( \mathbf{X} ) \circledast \mathbf{W}_c,
\label{eqn:STFNet_conv}
\end{equation}
where $\mathbf{SpectralPad}(\cdot)$ denotes our spectral padding operation, and $\circledast$ denotes the convolution operation.

Next, we discuss the way to share the kernel weight matrix $\mathbf{W}_c$ with multi-resolution data. Other than interpolating the kernel weight matrix as shown in~\eqref{eqn:linear_interpolate} and~\eqref{eqn:spectral_interpolate}, we propose another solution for the STFNet-convolution operation. The convolution operation concerns more about the pattern of relative positions on the frequency domain. 
Therefore, instead of providing additional kernel details on fine-grained frequency resolution, we can just ensure that the convolution kernel is applied with the same frequency spacing on representations with different frequency resolutions. 
Such idea can be implemented with the dilated convolution~\cite{yu2015multi}. 
If $\mathbf{W}_c$ is applied to a input tensor with $K=\lfloor \tau/2\rfloor +1$ frequency components, for a input tensor with $K'=\lfloor \tau'/2\rfloor +1$ frequency components ($\tau' > \tau$), the dilated rate $r$ is set to $\tau'/\tau - 1$.
A simple illustration of STFNet-convolution with dilated configuration is shown in Figure~\ref{fig:STFNet_convolution}.

\subsection{STFNet-Pooling Operation}~\label{sec:STFNet_pooling}
In order to provide a dimension reduction method for sensing series within STNet, we introduce the STFNet-pooling operation. STFNet-pooling truncates the spectral information over time with a pre-defined frequency pattern. As a widely-used processing technique, filtering zeroes unwanted frequency components in the signal. Various filtering techniques have been designed, including low-pass filtering, high-pass filtering, and band-pass filtering, which serve as templates for our STFNet-pooling. Instead of zeroing unwanted frequency components, STFNet-pooling removes unwanted components and then concatenates the left pieces. 
For applications with domain knowledge about signal-to-noise ratio over the frequency domain, specific pooling strategy can be designed. In this paper, we focus on low-pass STFNet-pooling as an illustrative example.

To extend the STFNet-pooling operation to multiple resolutions and preserving spectral information, we make sure that all representations have the same cut-off frequency according to their own frequency resolutions. A simple example of low-pass STFNet-pooling operation is shown in Figure~\ref{fig:STFNet_pooling}. We can see that our three tensors are truncated according to the same cut-off frequency, $f_s/4$.

%
%


\section{Evaluation}~\label{sec:evaluation}
In this section, we evaluate the STFNet with diverse sensing modularities. We focus on the device-based and device-free human activity recognitions with motion sensors (accelerometer and gyroscope), WiFi, ultrasound, and visible light. We first introduce the experimental setting, including data collection and baseline algorithms. Next, we show the performance metrics of leave-one-user-out evaluation of human activity recognition with different modularities. Finally, we analyze the effectiveness of STFNet through several ablation studies.

\subsection{Experimental Settings}~\label{sec:eval_setting}
In this subsection, we first introduce detailed information of the dataset we used or collected for each evaluation task. Then we specify the way to test the performance of evaluation task.

\textbf{\textit{Motion Sensor:}} In this experiment, we recognize human activity with motion sensors on smart devices.
We use the dataset collected by Allan et al.~\cite{stisen2015smart}. This dataset contains readings from two motion sensors (accelerometer and gyroscope). Readings were recorded when users executed activities scripted in no specific order, while carrying smartwatches and smartphones. The dataset contains 9 volunteers, 6 activities (biking, sitting, standing, walking, climbStair-up, and climbStair-down). 
We align two sensor readings, linear interpolate two readings by 100Hz, and segment them into non-overlapping data samples with time interval 5.12s. 
Therefore, each data sample is a $512\times 6$ matrix, where both accelerometer and gyroscope have readings on $x$, $y$, and $z$ axis.

\textbf{\textit{WiFi:}} In this experiment, we make use of Channel State Information (CSI) to analyze human activities. CSI refers to the known channel properties of a communication link, which can be affected by the presence of humans and their activities. We employ 11 volunteers (including both men and women) as the subjects and collect CSI data from 6 different rooms in two different buildings.
In particular, we build a WiFi infrastructure, which includes a transmitter (a wireless router) and two receivers. 
We use the tool to report CSI values of 30 OFDM subcarriers~\cite{halperin2011tool}. 
The experiment contains 6 activities (wiping the whiteboard, walking, moving a suitcase, rotating the chair. sitting, as well as standing up and sitting down).
We linearly interpolate the CSI data with a uniform sampling period, and down-sample the measurements into 100Hz. Then we segment the down-sampled CSI data into non-overlapping data samples with time interval 5.12s. Therefore, each data sample is a $512\times 30$ matrix, where each CSI measurement has readings from 30 subcarriers.

\textbf{\textit{Ultrasound:}} In this experiment, we conduct human activity recognition based on ultrasound. We employ 12 volunteers as the subjects to conduct the 6 different activities. The activity data are collected from 6 different rooms in two different buildings.  The transmitter is an iPad on which an ultrasound generator app is installed, and it can emit an ultrasound signal of approximately 19 KHz. The receiver is a smartphone and we use the installed recorder app to collect the sound waves. We demodulate the received signal with carrier frequency 19KHz, and down-sample the measurement into 100Hz. Then we segment the down-sampled ultrasound data into non-overlapping data samples with time interval 5.12s. Therefore, each sample is a $512\times1$ matrix.

\textbf{\textit{Visible light:}} In this experiment, we capture the human activity in the visible light system. We build an optical system using photoresistors to capture the in-air body gesture, which can detect the illuminance change (lux) caused by the body interaction. 
In the experiment, there are three light conditions (natural mode, warm mode, and cool mode) and 4 hand gestures (drawing an anticlockwise circle, drawing a clockwise circle, drawing a cross, and shaking hand side to side). 
We employ 6 volunteers as the subjects and each of them performs 20 trials of every gesture under a given lighting condition. We linearly interpolate and down-sample the measurements into 25Hz. Then we segment the data into non-overlapping data samples with time interval 5.12s. Therefore, each sample is a $128\times6$ matrix, where each measurement contains readings from 6 CdS cells.

\begin{table}[!htp]
\vspace{-0.3cm}
\begin{center}
\caption {Illustration of models with two sensor inputs.}
\vspace{-0.3cm}
\label{tab:exp_model}
\scriptsize
\begin{tabular}{| c | c | c | c |}
\hline
 \multicolumn{2}{ |c| }{STFNet-Filter/Conv} & \multicolumn{2}{ c| }{DeepSense/ComplexNet}\\
\hline
\hline
 \multirow{ 2}{*}{Sensor Data 1} & \multirow{ 2}{*}{Sensor Data 2} &  Chunked &  Chunked \\
 & & Sensor Data 1 & Sensor Data 2 \\
 \hline
 STFNet1-1 & STFNet1-2 & Conv Layer1-1 & Conv Layer1-2\\
 \hline
  STFNet2-1 & STFNet2-2 & Conv Layer2-1 & Conv Layer2-2\\
 \hline
  STFNet3-1 & STFNet3-2 & Conv Layer3-1 & Conv Layer3-2\\
 \hline
  \multicolumn{2}{ |c| }{STFNet-pooling} & \multicolumn{2}{ c| }{Max pooling}\\
\hline
 \multicolumn{2}{ |c| }{STFNet4} & \multicolumn{2}{ c| }{Conv Layer4}\\
\hline
 \multicolumn{2}{ |c| }{STFNet5} & \multicolumn{2}{ c| }{Conv Layer5}\\
\hline
 \multicolumn{2}{ |c| }{STFNet6} & \multicolumn{2}{ c| }{Conv Layer6}\\
\hline
\multicolumn{2}{ |c| }{Averaging} & \multicolumn{2}{ c| }{GRU}\\
\hline
\multicolumn{2}{ |c| }{Softmax} & \multicolumn{2}{ c| }{Softmax}\\
\hline
\end{tabular}
\end{center}
\end{table}
\setlength{\textfloatsep}{0pt}

\textbf{\textit{Testing:}} In the whole evaluation, to illustrate the generalization ability of STFNet and other baseline models, we perform leave-one-user-out cross validation for every task. For each time, we choose the data from one user as testing data with the left as training data.  We then compare the performance of models according to their accuracy and F1 score with $95\%$ confidence interval.

\subsection{Models in Comparison}~\label{sec:baseline}
In order to evaluate, when compared to conventional deep learning components (\ie convolutional and recurrent layers), whether our proposed STFNet component is better at decoding information and extracting features from sensing inputs, we substitute  components in the state-of-the-art neural network structure for IoT applications with STFNet. In the whole evaluation, we choose DeepSense as the state-of-the-art structure, which has shown signifiant improvements on various sensing tasks~\cite{yao2017deepsense}. The illustration of structures of five comparing models with two sensor inputs are shown in Table~\ref{tab:exp_model}. Detailed information of our comparing models are listed as follows,

\begin{enumerate}
\item \textbf{\textit{STFNet-Filter:}} This model integrates the proposed STFNet component and the DeepSense structure. Within the STFNet component, we use the STFNet-filtering operation designed in Section~\ref{sec:STFNet_filtering}. The intuition of DeepSense structure is to first perform local processing within each sensor and then perform global sensor fusion over multiple sensors. In this model, we replace all convolutional layers used in local/global sensor data processing with our time-frequency analyzing component, STFNet. Since our model has already incorporated time-domain analysis within the STFNet component through multi-resolution processing, we replace the Gated Recurrent Units (GRU) with simple feature averaging time at last.
\item \textbf{\textit{STFNet-Conv:}} This model is almost the same as the STFNet-Filter, except that we use the STFNet-convolution operation designed in Section~\ref{sec:STFNet_convolution}. 
\item \textbf{\textit{DeepSense-Freq:}} This model is the original DeepSense~\cite{yao2017deepsense}. It divides the input sensing data into chunks, and processes each chunk with DFT. It treats the real and imagery parts of discrete Fourier transformed time chunks as the additional feature dimensions. This is the state-of-the-art deep learning model for sensing data modelling and IoT applications.
\begin{figure}[!htb]
\vspace{-0.5cm}
\centering
\includegraphics[width=0.675\linewidth]{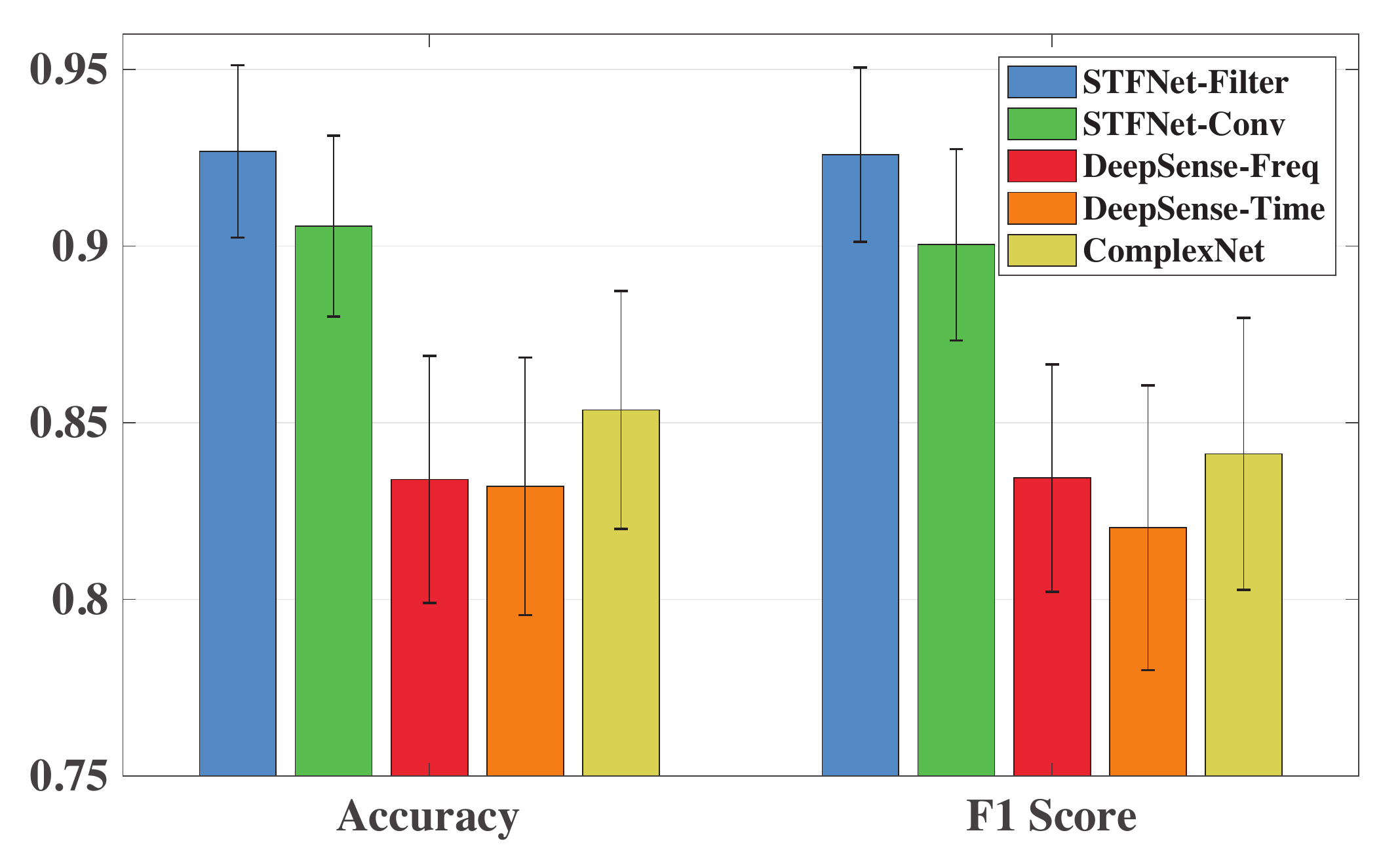}
\vspace{-0.3cm}
\caption{The accuracy and F1 score with $95\%$ confidence interval for motion sensors.}
\label{fig:HHAR_acc_f1}
\vspace{-0.4cm}
\end{figure}
\vspace{-0.1cm}
\begin{figure}[!htb]
\centering
\includegraphics[width=0.675\linewidth]{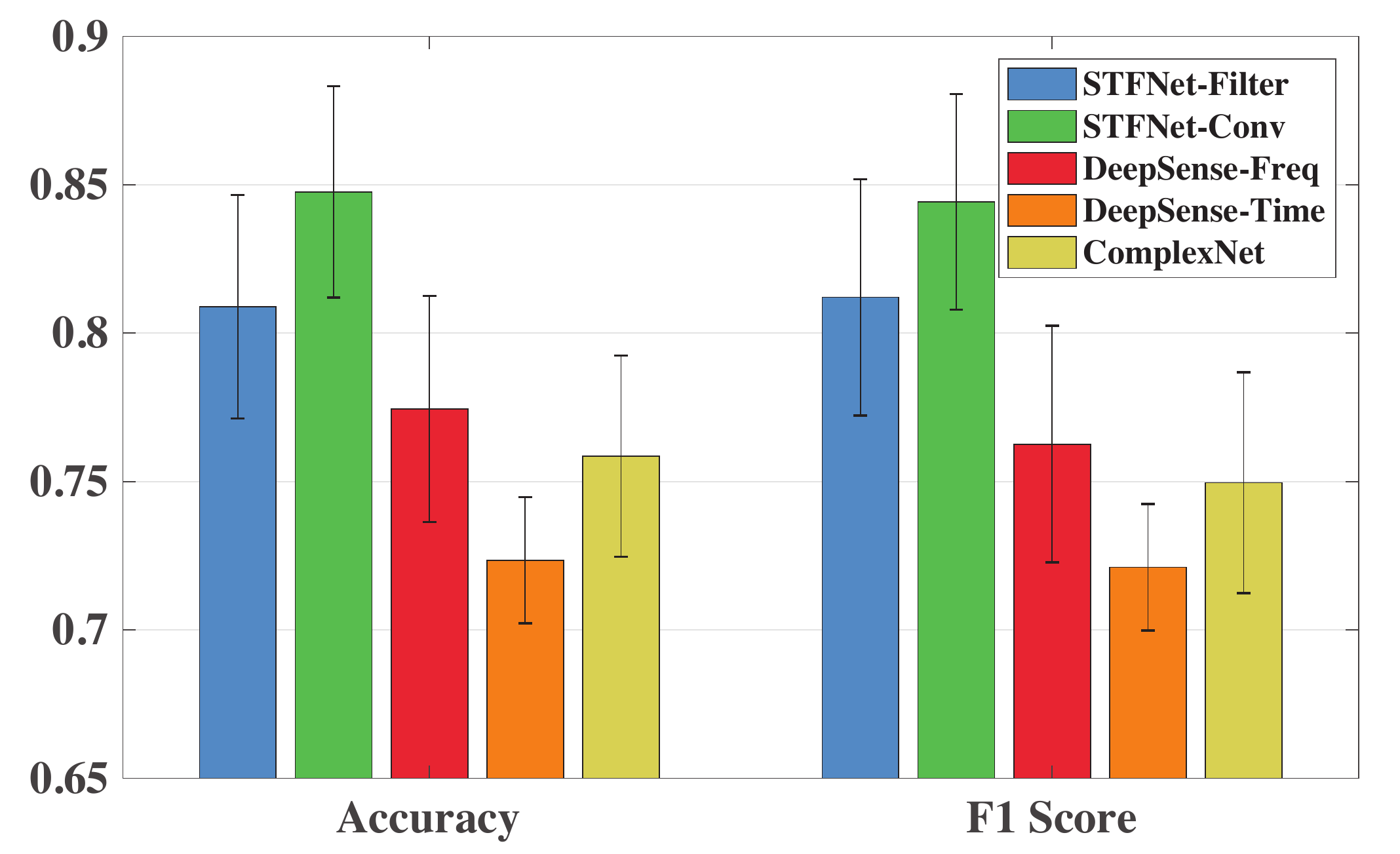}
\vspace{-0.4cm}
\caption{The accuracy and F1 score with $95\%$ confidence interval for  WiFi.}
\label{fig:WIFI_acc_f1}
\end{figure}
\item \textbf{\textit{DeepSense-Time:}} This model is almost the same as the DeepSense-Freq, except that it directly takes the chunked raw sensing data without DFT as input.
\item \textbf{\textit{ComplexNet:}} This model is a complex-value neural network~\cite{trabelsi2017deep} that can operate on complex-value inputs.
Instead of using simple CNN and RNN structure as originally proposed~\cite{trabelsi2017deep}, we cheat in their favor by using the DeepSense structure, which improves the performance in all tasks.
The network inputs are chunked sensing data with DFT. 
\end{enumerate}

\subsection{Effectiveness}
In this section, we discuss about the effectiveness of our proposed STFNet based on extensive experiments and diverse sensing modalities, compared with other state-of-the-art deep learning models.

As we mentioned in Section~\ref{sec:eval_setting}, all models are evaluated through leave-one-user-out cross validation with accuracy and F1 score accompanied by the $95\%$ confidence interval. STFNet-based models (STFNet-Filter and STFNet-Conv) take a sliding window set for multi-resolution short-time Fourier transform. 
We choose the set to be $\{16, 32, 64, 128\}$ for activity recognition with motion sensors, WiFi, and ultrasound; and choose set to be $\{8, 16, 32, 64\}$ for activity recognition with visible light. DeepSense-based models (DeepSense-Freq and DeepSense-Time) need a sliding window for chunking input signals. In the evaluation, we cheat in their favor by choosing the best-performing window size from $\{8, 16, 32, 64, 128\}$ according to the accuracy metric. 
In addition, we consistently configure STFNet-filtering operation with linear interpolation, and STFNet-convolution operation with spectral padding. We will show further evaluations on multi-resolution operations and the effects of diverse operation settings in Section~\ref{sec:ablation}.

\subsubsection{Motion Sensors}
For device-based activity recognition with motion sensors, there are 9 users. The accuracy and F1 score with the $95\%$ confidence interval for leave-one-user-out cross validation are illustrated in Figure~\ref{fig:HHAR_acc_f1}.
\begin{figure}[!htb]
\vspace{-0.5cm}
\centering
\includegraphics[width=0.675\linewidth]{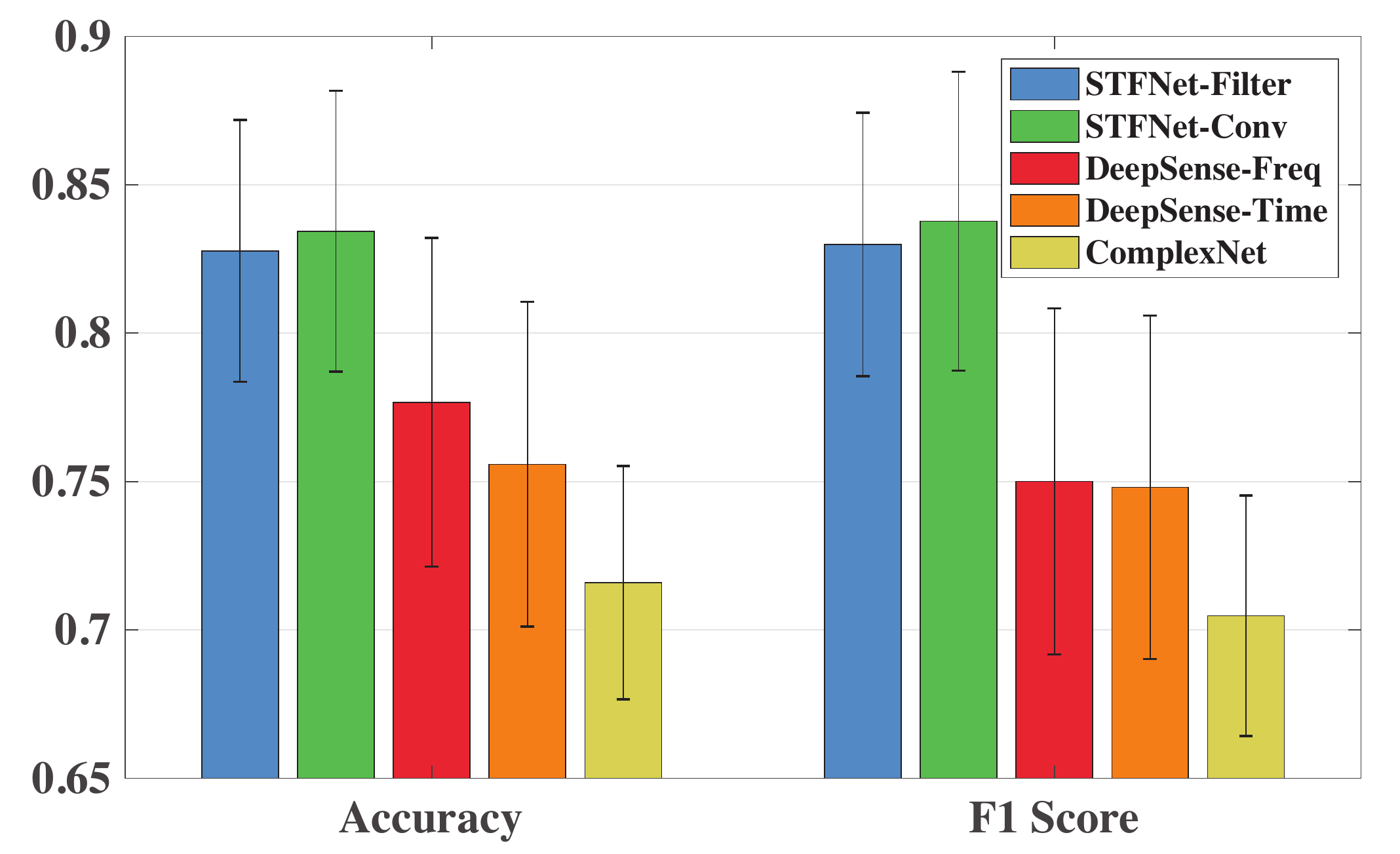}
\vspace{-0.3cm}
\caption{The accuracy and F1 score with $95\%$ confidence interval for Ultrasound.}
\label{fig:acoustic_acc_f1}
\end{figure}
\begin{figure}[!htb]
\vspace{-0.4cm}
\centering
\includegraphics[width=0.675\linewidth]{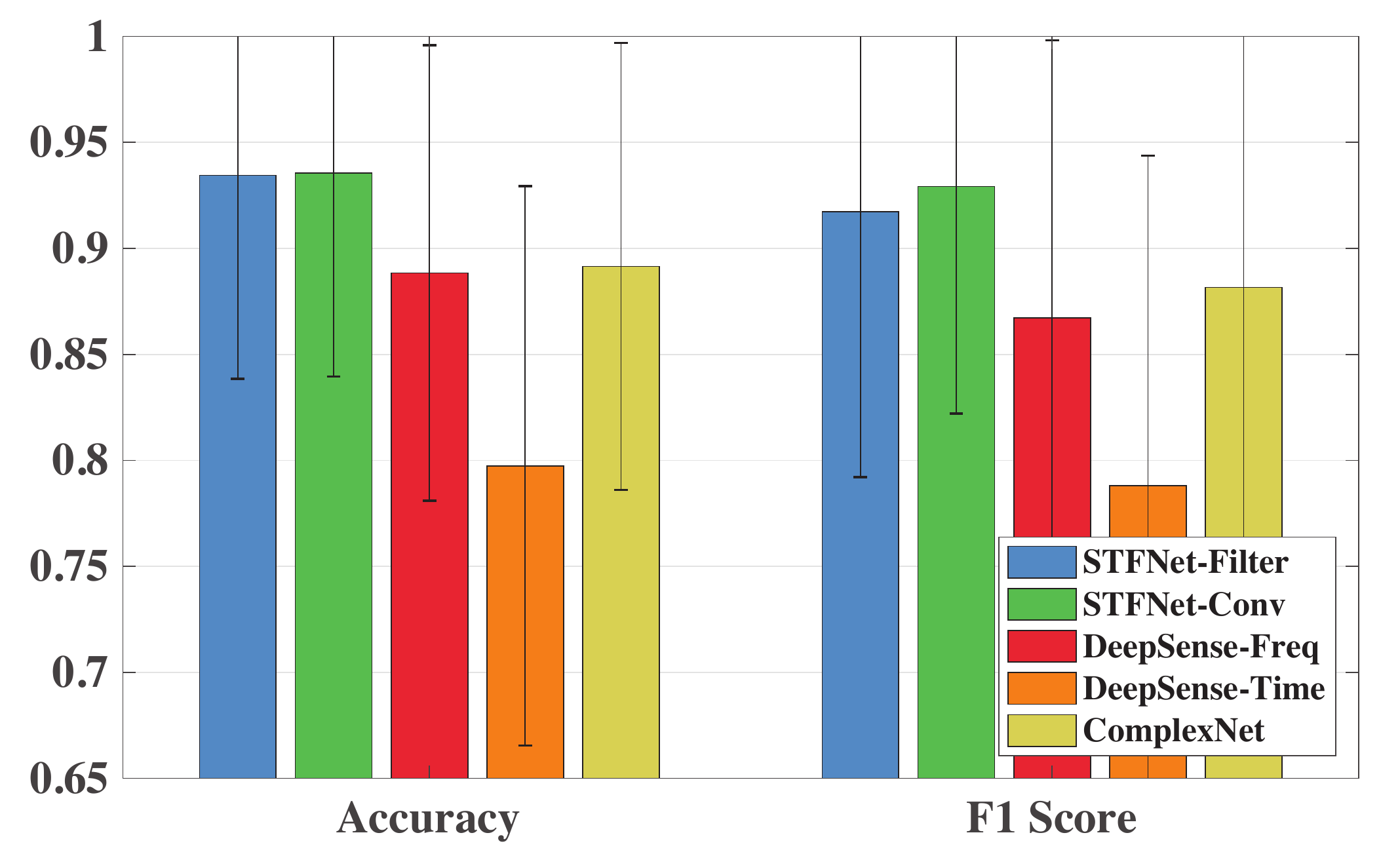}
\vspace{-0.3cm}
\caption{The accuracy and F1 score with $95\%$ confidence interval for Visible light.}
\label{fig:light_acc_f1}
\end{figure}
STFNet based models, \ie STFNet-Filter and STFNet-Conv, outperform all other baseline models with a large margin. The confidence interval lower bound of STNet-Filter and STFNet-Conv is even better than the confidence interval upper bound of DeepSense-Freq and DeepSense-Time. 
STFNet-Filter performs better than STFNet-Conv in this experiment, indicating that different activities have distinct global profiling patterns with motion sensor readings in the frequency domain, even among different users. STFNet-Filter is able to learn the accurate global frequency profiling, which makes it the top-performance model in this task. In addition, compared to ComplexNet, STFNet based models show clear improvements. Therefore, using just complex-value neural network for sensing signal is far from enough. The multi-resolution processing and operations that are spectral-compatible are all crucial designs.

\subsubsection{WiFi}
For device-free activity recognition with WiFi signal, there are 11 users. The accuracy and F1 score with the $95\%$ confidence interval for leave-one-user-out cross validation are illustrated in Figure~\ref{fig:WIFI_acc_f1}.
STFNet based models still outperform all others with a clear margin, illustrating the effectiveness of principled design of STFNet from time-frequency perspective. DeepSense-Freq outperforms DeepSense-Time in this experiment, which means that even having time-frequency transformation as pre-processing can help. The complex-value network, ComplexNet, performs worse than its real-value counterpart, DeepSense-Freq. This indicates that blindly processing time-frequency representations without preserving their physical meanings can even hurt the final performance. STNet-Conv performs better than STNet-Filter in the WiFi experiment, indicating that local shiftings in the frequency domain are more representative for diverse activities profiled with WiFi CSI.

\begin{figure}[!htb]
\vspace{-0.5cm}
\begin{subfigure}{\linewidth}
  \centering
  \includegraphics[width=0.675\linewidth]{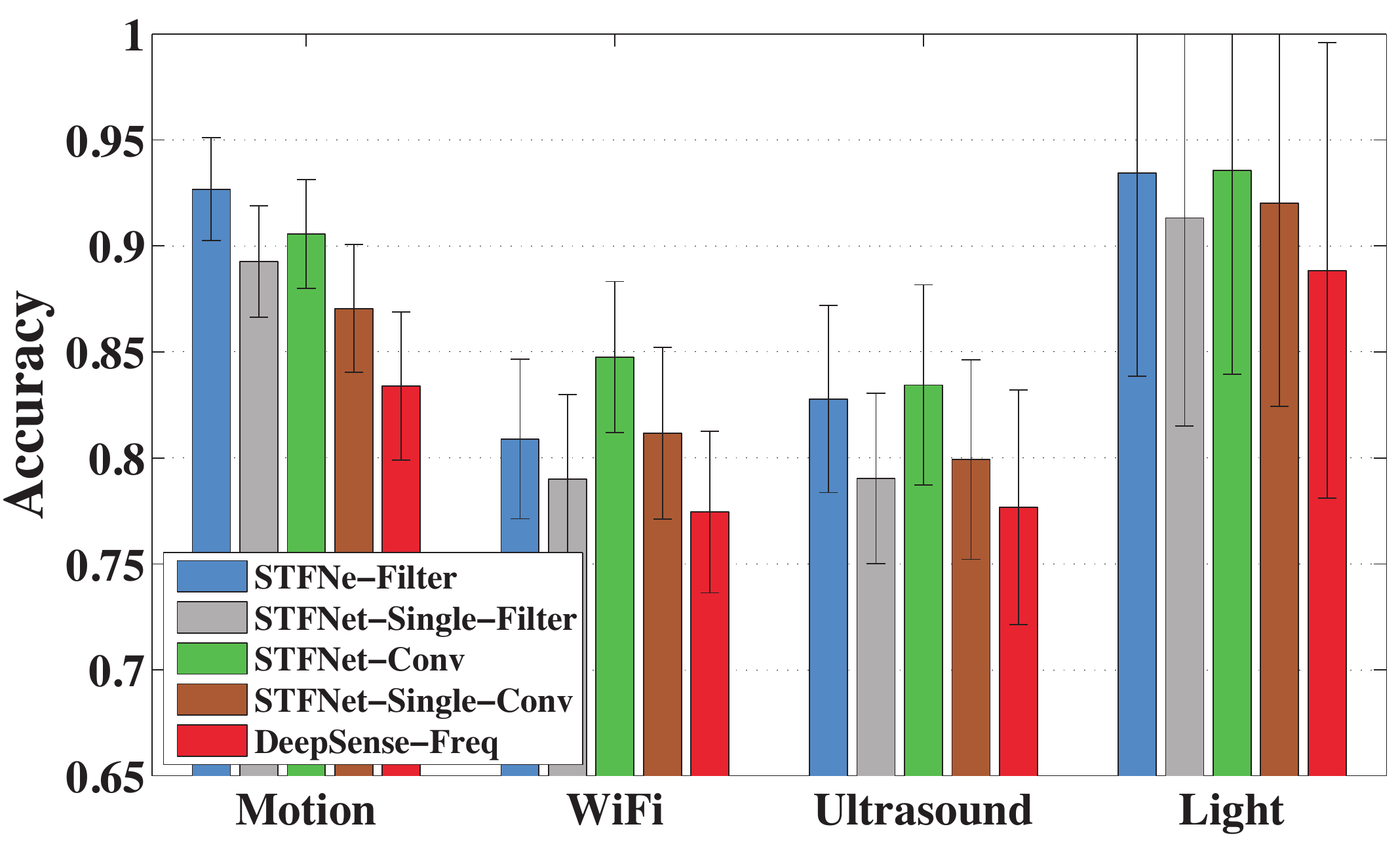}
  \vspace{-0.3cm}
  \caption{Accuracy with $95\%$ confidence interval. }
  \label{fig:ablation_multi_single_acc}
\end{subfigure}%
\vspace{-0.1cm}
\begin{subfigure}{\linewidth}
  \centering
  \includegraphics[width=0.675\linewidth]{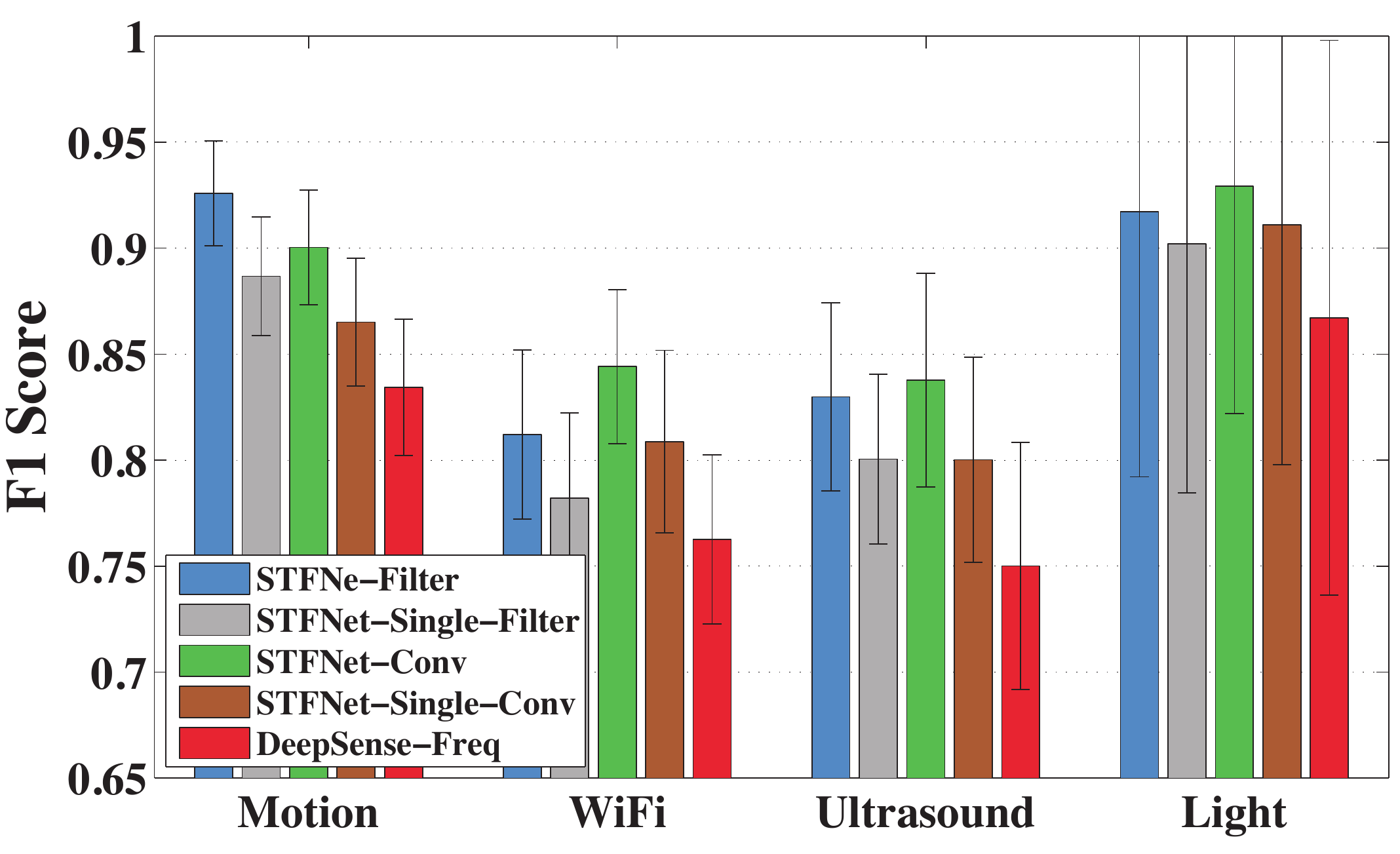}
  \vspace{-0.3cm}
  \caption{F1 score with $95\%$ confidence interval.}
  \label{fig:ablation_multi_single_f1}
\end{subfigure}
\vspace{-0.3cm}
\caption{Multi-Resolution v.s. Single-Resolution}
\label{fig:ablation_multi_single}
\end{figure}

\subsubsection{Ultrasound}
There are 12 users in device-free activity recognition with ultrasound experiment. The accuracy and F1 score with the $95\%$ confidence interval for leave-one-user-out cross validation are illustrated in Figure~\ref{fig:acoustic_acc_f1}. 
STFNet based models still significantly outperforms all other baselines. An interesting observation is that ComplexNet performs even worse than both DeepSense-Freq and DeepSene-Time, which again validates the importance of designing neural networks for sensing signal with multi-resolution processing as well as preserving the time and frequency information.

\subsubsection{Visible Light}
There are 6 users in the experiment of device-free activity recognition with visible light. The accuracy and F1 score with the $95\%$ confidence interval are illustrated in Figure~\ref{fig:light_acc_f1}. Except for the DeepSense-Time, all other models can can achieve an accuracy of approximately 90\% or higher. STFNet based models still do the best. There is no significant  difference between STFNet-Filter and STFNet-Conv, which indicates that measured visible light readings have quite clean representations in the frequency domain.

\subsection{Ablation Studies}~\label{sec:ablation}
In the previous section, we illustrate the performance of STFNet compared to other state-of-the-art baselines. In this section, we focus mainly on the STFNet design. We conduct several ablation studies by deleting one designing feature from STFNet at a time.

\subsubsection{Multi-Resolution v.s. Single-Resolution}
First, we validate the effectiveness of our design of multi-resolution processing in STFNet block. As shown in Figure~\ref{fig:STFNet_overview}, this includes multi-resolution STFT, hologram interleaving, and weights sharing techniques in STFNet-Filtering and STFNet-Convolution operations. In this experiment, we add two more baseline models, STFNet-Single-Filter and STFNet-Single-Conv, generated by deleting the multi-resolution processing in STFNet-Filter and STFNet-Conv respectively. These two models pick the best-performing window size from $\{8, 16, 32, 64, 128\}$ according to the accuracy metric. The results for all four tasks are illustrated in Figure~\ref{fig:ablation_multi_single}, where DeepSense-Freq severs as a decent performance low-bound. The design of multi-resolution processing significantly impacts the performance of STFNet. STFNet-Single-Filter and STFNet-Single-Conv show clear performance degradation compared to their multi-resolution counterparts. In addition, STFNet-Single-Filter and STFNet-Single-Conv still consistently outperform DeepSense-Freq with a clear margin. This is because our other designed operations, including STFNet-Filtering, STFNet-Convolution, STFNet-Pooling still facilitate the learning in time-frequency domain.

\begin{figure}[!htb]
\vspace{-0.5cm}
\begin{subfigure}{\linewidth}
  \centering
  \includegraphics[width=0.675\linewidth]{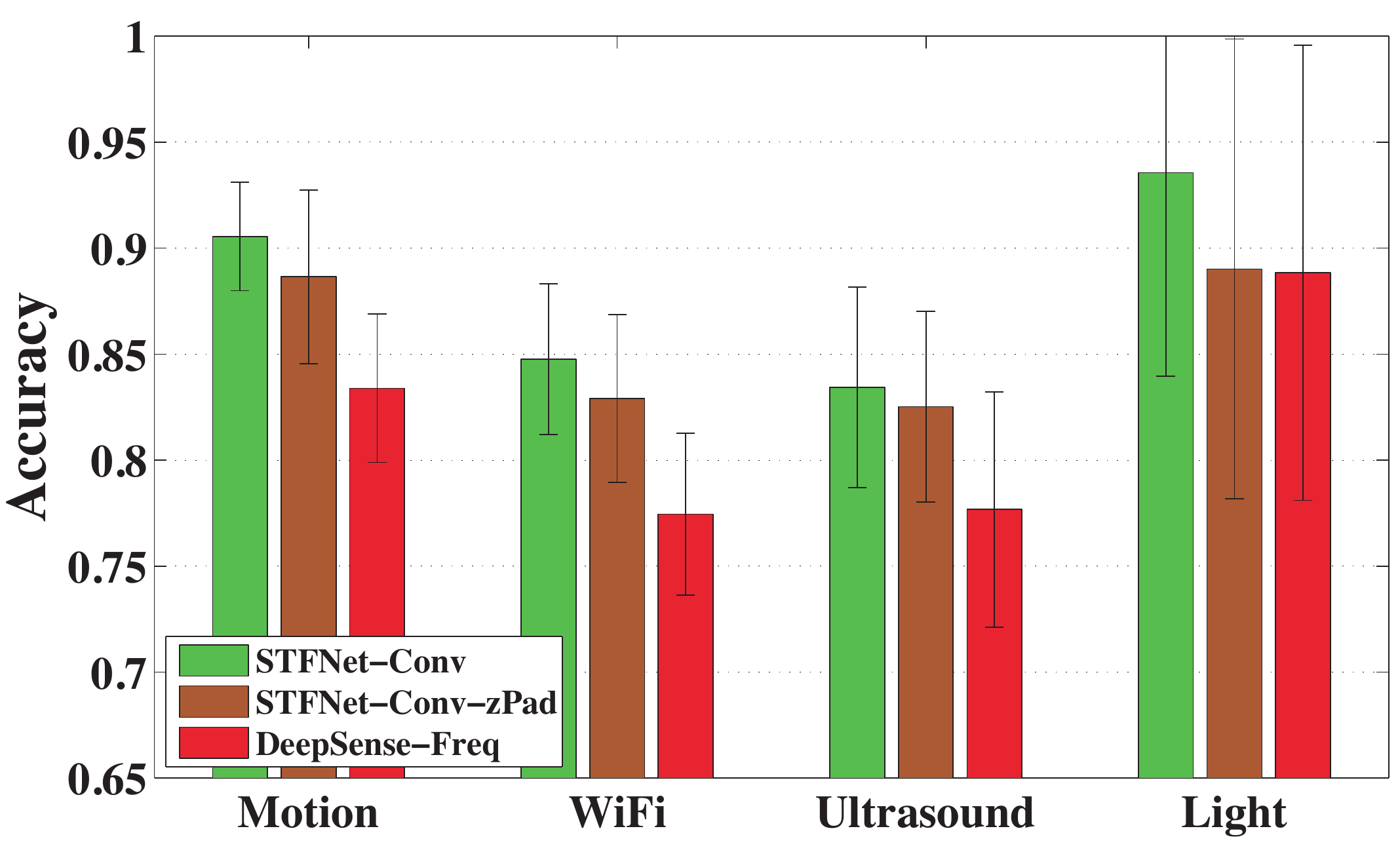}
  \vspace{-0.3cm}
  \caption{Accuracy with $95\%$ confidence interval. }
  \label{fig:ablation_padding_acc}
\end{subfigure}%
\vspace{-0.1cm}
\begin{subfigure}{\linewidth}
  \centering
  \includegraphics[width=0.675\linewidth]{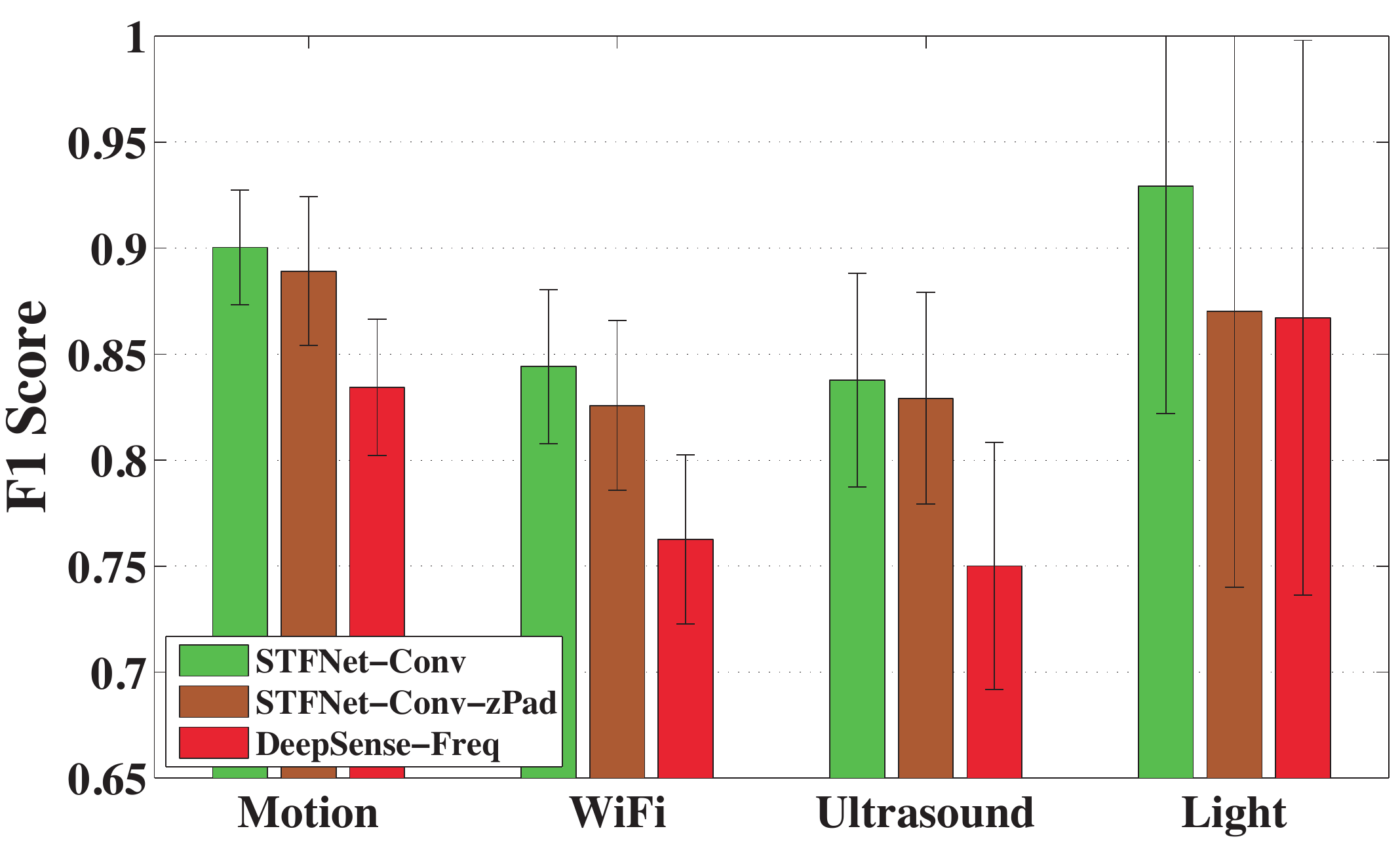}
  \vspace{-0.35cm}
  \caption{F1 score with $95\%$ confidence interval.}
  \label{fig:ablation_padding_f1}
\end{subfigure}
\vspace{-0.4cm}
\caption{Spectral Padding v.s. Zero Padding}
\label{fig:ablation_padding}
\end{figure}

\subsubsection{Spectral Padding v.s. Zero Padding} Next, we validate our design of spectral padding in the STFNet-Convolution operation as shown in Figure~\ref{fig:STFNet_convolution}. In this experiment, we add a new baseline algorithm, STFNet-Conv-zPad, by replacing spectral padding with traditional zero padding in the STFNet-Conv. The accuracy and F1 score of all four tasks are shown in Figure~\ref{fig:ablation_padding}. Here, DeepSense-Freq is still treated as a performance low-bound. By comparing STFNet-Conv-zPad and STFNet-Conv, we can see that spectral padding consistently helps improving the model performance. In most cases, the improvement is limited. However, in the case of visible light, spectral padding significantly improves both accuracy and F1 score. Therefore, designing neural network by preserving the time-frequency semantics of sensing signal is an important rule to follow.

\vspace{-0.2cm}
\subsubsection{Linear Interpolation v.s. Spectral Interpolation} Then, we compare our two designs of weight interpolation method in the STFNet-Filtering operation, linear interpolation and spectral interpolation, as shown in Figure~\ref{fig:STFNet_filtering}. The STFNet-Filter defined in Section~\ref{sec:baseline} uses linear interpolation, so we rename it as STFNet-Filter-LinearInpt in this experiment. We add a new baseline model called STFNet-Filter-SpectralInpt by using spectral interpolation instead of linear interpolation in STFNet-Filter. The results of all four tasks are illustrated in Figure~\ref{fig:ablation_interp}. In general, the performance of two design choices are almost the same. At most of time, linear interpolation performs slightly better. In addition, we recommend using linear interpolation, since its implementation is easier,

\begin{figure}[!htb]
\vspace{-0.5cm}
\begin{subfigure}{\linewidth}
  \centering
  \includegraphics[width=0.675\linewidth]{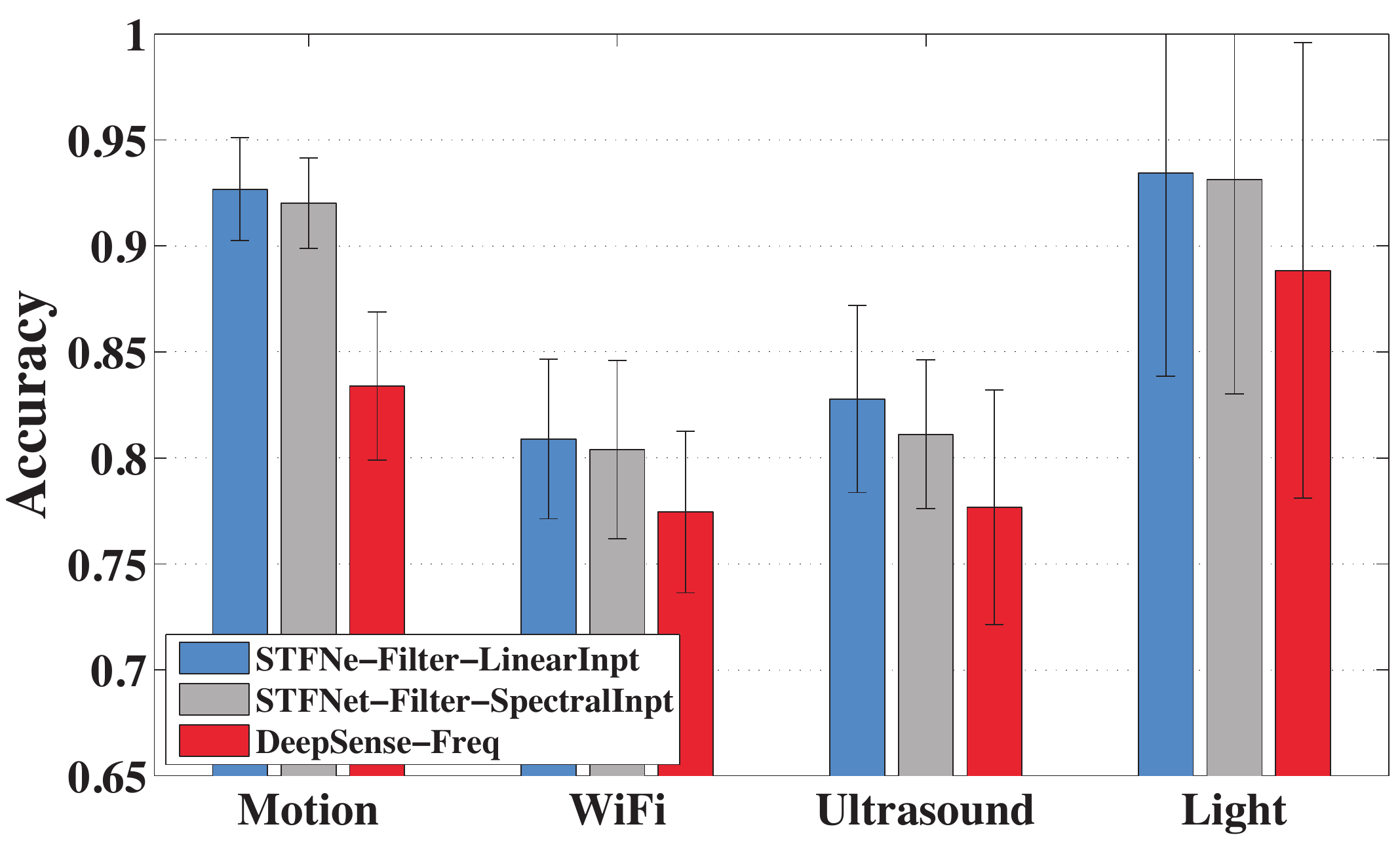}
  \vspace{-0.3cm}
  \caption{Accuracy with $95\%$ confidence interval. }
  \label{fig:ablation_interp_acc}
\end{subfigure}%
\vspace{-0.1cm}
\begin{subfigure}{\linewidth}
  \centering
  \includegraphics[width=0.675\linewidth]{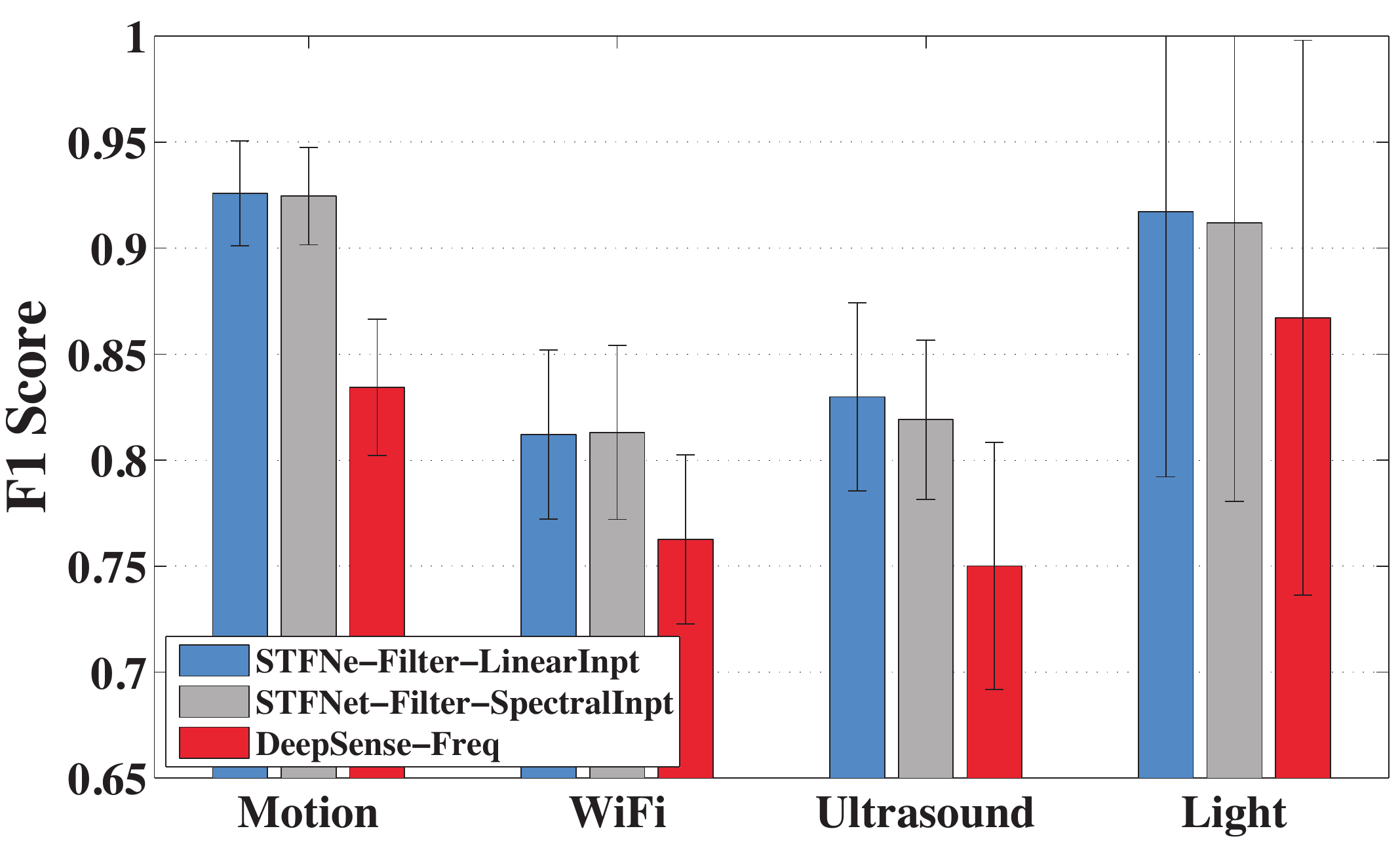}
  \vspace{-0.3cm}
  \caption{F1 score with $95\%$ confidence interval.}
  \label{fig:ablation_interp_f1}
\end{subfigure}
\vspace{-0.4cm}
\caption{Linear Interpolation v.s. Spectral Interpolation}
\label{fig:ablation_interp}
\end{figure}

\vspace{-0.2cm}
\subsubsection{STFNet Pooling v.s. Mean/Max Pooling} Finally, we validate our design of STFNet-Pooling (low-pass deisgn) as shown in Figure~\ref{fig:STFNet_pooling}. In this experiment, we add two new baseline algorithms, STFNet-Filter-mPad and STFNet-Conv-mPad, by replacing STFNet-Pooling in STFNet-Filter and STFNet-Conv with traditional max/mean pooling in the time domain (through choosing the one has better accuracy). The results are illustrated in Figure~\ref{fig:ablation_pool}. In all settings, STFNet-Pooling shows better performance. In some cases, the improvement are significant. We believe that STFNet-Pooling can achieve even better performance if given the detailed signal-to-noise ratio over the frequency domain for each specific sensor. Then we can employ other pooling strategies instead of the low-pass design.

\section{Discussion}~\label{sec:discussion}
This paper provides a principled way of designing neural networks for sensing signals inspired by the fundamental nature of the underlying physical processes. STFNet, operates directly in the frequency domain, in which the measured physical phenomena are best exposed. We propose three types of learnable frequency manipulations that are able to operate on multi-resolution representations, while preserving the underlying time-frequency information. Although extensive experiments have illustrated the superior performance of STFNet, 
further research is needed to better understand design choices for neural networks from the time-frequency perspective.

One challenge is to explore the possibility of integrating neural networks with other time-frequency transformations. In this paper, STFNet focuses on the short-time Fourier Transform. However, STFT is the most basic one. There are plenty of other transformation basis functions in traditional time-frequency analysis. How to naturally integrate them with neural network while keeping the underlying physical meaning within transformed representations? How to choose or design the most suitable transformation basis functions that meet the corresponding mathematical requirements? Answers to these questions can greatly impact the way researchers design neural networks for sensing signal processing.
\begin{figure}[!htb]
\vspace{-0.5cm}
\begin{subfigure}{\linewidth}
  \centering
  \includegraphics[width=0.675\linewidth]{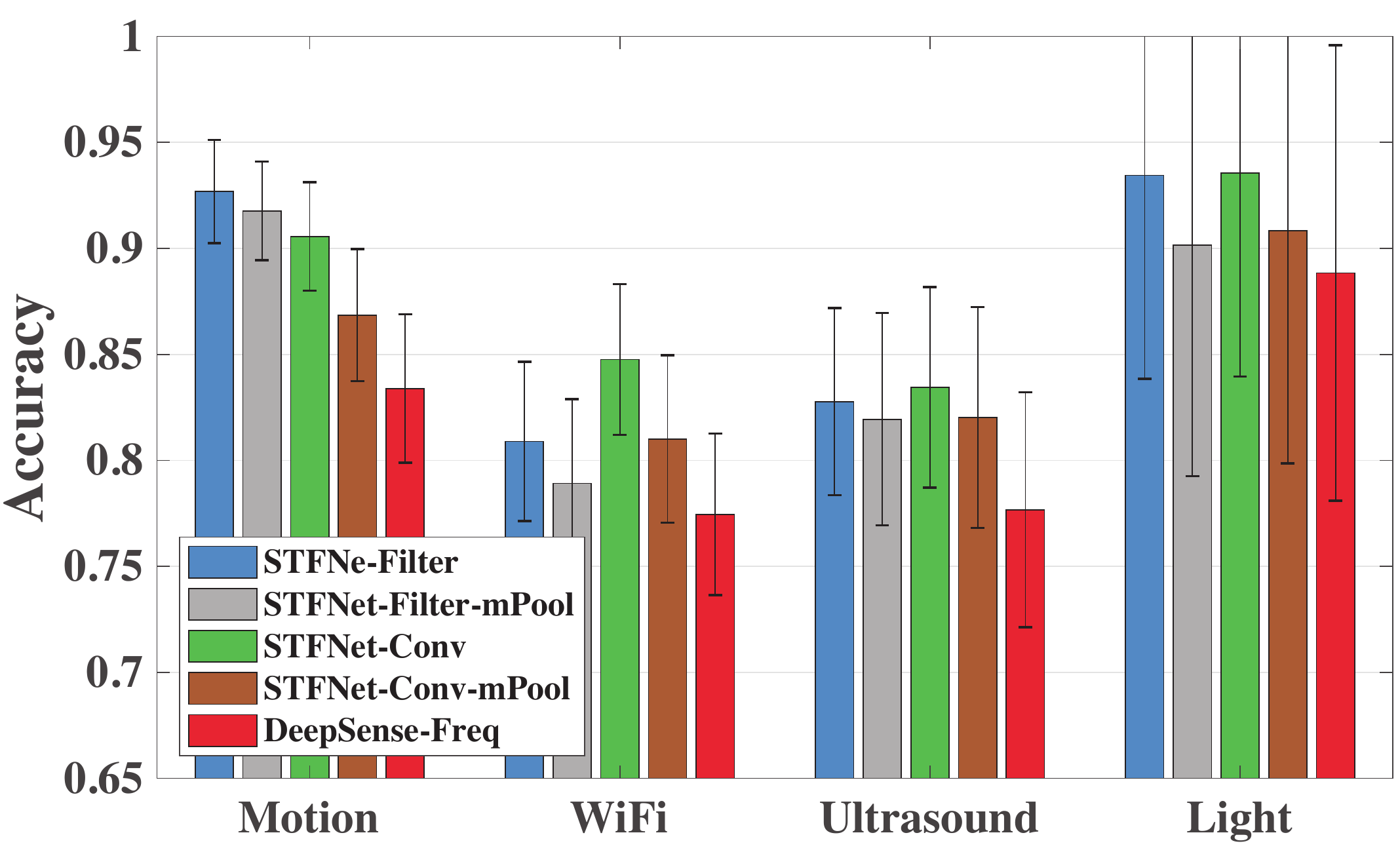}
  \vspace{-0.3cm}
  \caption{Accuracy with $95\%$ confidence interval. }
  \label{fig:ablation_pool_acc}
\end{subfigure}%
\vspace{-0.1cm}
\begin{subfigure}{\linewidth}
  \centering
  \includegraphics[width=0.675\linewidth]{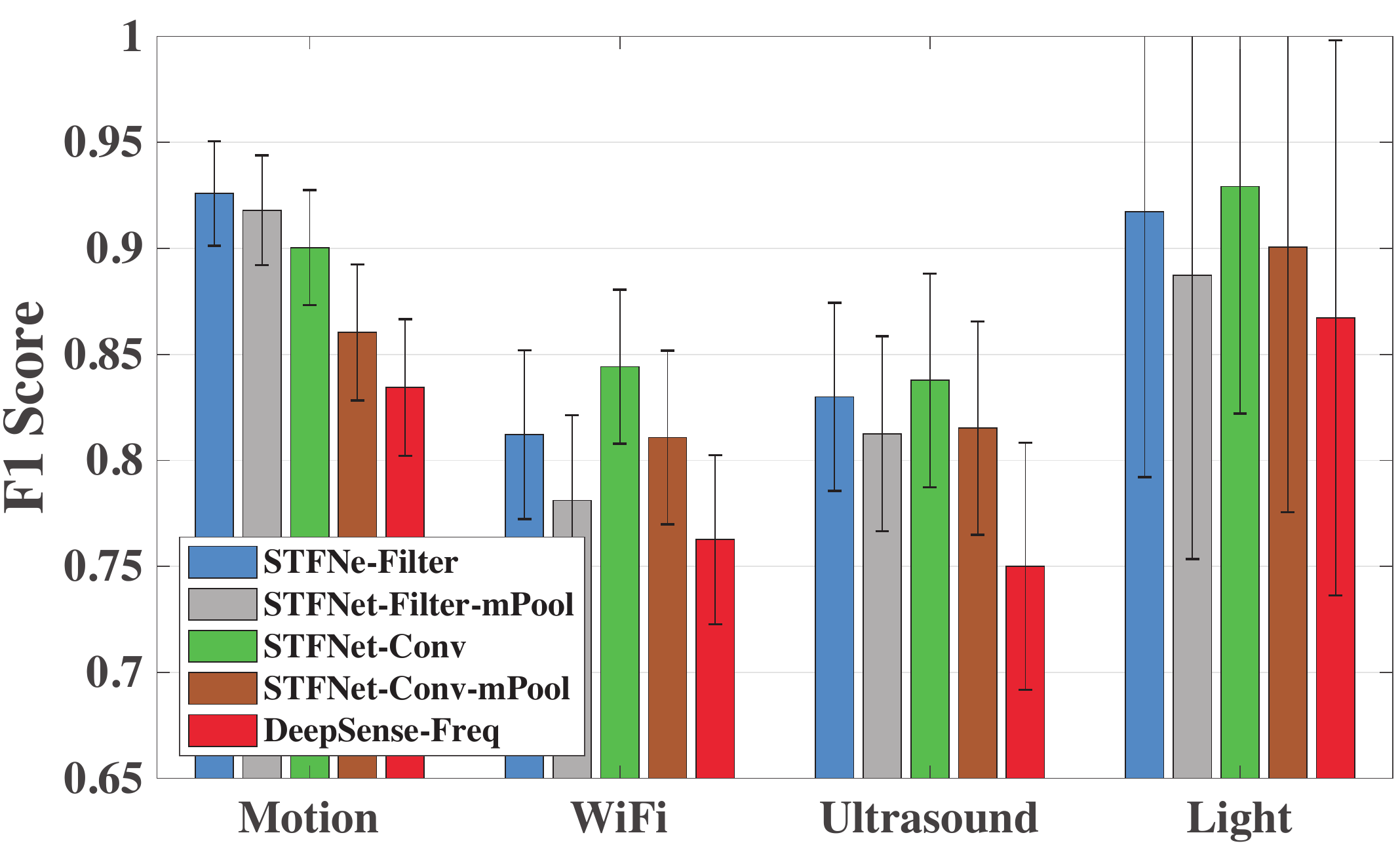}
  \vspace{-0.3cm}
  \caption{F1 score with $95\%$ confidence interval.}
  \label{fig:ablation_pool_f1}
\end{subfigure}
\vspace{-0.4cm}
\caption{STFNet Pooling v.s. Mean/Max Pooling}
\label{fig:ablation_pool}
\vspace{-0.3cm}
\end{figure}
Another challenge is to empower the frequency manipulations to have heterogeneous behaviours over the time. In STFNet, all designed operations are learnable frequency manipulations, which perform identically over time. In order to fully exploit the potential of time-frequency analysis, further research is needed on designing time-varying time-frequency manipulations, that adapt to current temporal patterns. 

Furthermore, a better experimental and theoretical understanding is needed of the basic settings of neural networks to support computation in time-frequency domain. For traditional real-value neural networks, researchers have good intuitions about the basic configurations of initialization, activation functions, dropout and normalization techniques, and optimization methods. However, for neural network in the time-frequency domain, our understanding is limited. Although the reseach community started to study the basic settings of neural networks with complex values~\cite{trabelsi2017deep}, the current understanding remains preliminary. Time-frequency analysis can have operations in both the real and complex domains. At the same time, the underlying time-frequency information within the internal representations can make the related studies even more complicated. We believe that this understanding will greatly facilitate future design of deep learning systems for IoT.

In addition, outside the IoT context, there exists a large number of transformations and dimension reduction techniques, such as SVD and PCA, that have made great impact in revealing useful features of complex phenomena. Our study of deep learning with STFT suggests that integrating deep neural networks with other common transformations may facilitate learning in domains where such transformations reveal essential features of the input signal domain. Future work is needed to explore this conjecture.

\vspace{-0.2cm}
\section{Conclusion}~\label{sec:conclusion}
In this paper, we introduced STFNet, a principled way of designing neural networks from the time-frequency perspective. STFNet endows time-frequency analysis with additional flexibility and capability. In addition to just parameterizing the frequency manipulations with deep neural networks, we bring two key insights into the design of STFNet. On one hand, STFNet leverages and preserves the frequency domain semantics that encode time and frequency information. On the other hand, STFNet circumvents the uncertainty principle through multi-resolution transform and processing. Evaluations show that STFNet consistently outperforms the state-of-the-art deep learning models with a clear margin under diverse sensing modalities, and our two designing insights significantly contribute to the improvement. The designs and evaluations of STFNet unveil the benefits of incorporating domain-specific modeling and transformation techniques into neural network design.

\vspace{-0.2cm}
\begin{acks}
Research reported in this paper was sponsored in part by NSF under grants CNS 16-18627 and CNS 13-20209 and in part by the Army Research Laboratory under Coop- erative Agreements W911NF-09-2-0053 and W911NF-17-2-0196. The views and conclusions contained in this document are those of the authors and should not be interpreted as representing the official policies, either expressed or implied, of the Army Research Laboratory, NSF, or the U.S. Govern- ment. The U.S. Government is authorized to reproduce and distribute reprints for Government purposes notwithstanding any copyright notation here on.
\end{acks}

}

\newpage
\bibliographystyle{ACM-Reference-Format}
\bibliography{reference}

\end{document}